\documentclass{article}
\PassOptionsToPackage{numbers, compress}{natbib}
\usepackage[preprint]{neurips_2024}

\usepackage[utf8]{inputenc} 
\usepackage[T1]{fontenc}    
\usepackage{url}            
\usepackage{booktabs}       
\usepackage{amsfonts}       
\usepackage{nicefrac}       
\usepackage{microtype}      
\usepackage{xcolor}         

\usepackage{mathtools}
\usepackage{amsmath}
\usepackage{amssymb}
\usepackage{xspace}
\usepackage{graphicx}
\usepackage{caption}
\usepackage{subcaption}
\usepackage{hyperref}       
\usepackage{algorithm, algpseudocodex}

\newcommand{\rspace}[1]{\mathbb{R}^{\mathit{#1}}}
\newcommand{\bigo}[1]{\mathit{O(#1)}\xspace}
\newcommand{\set}[1]{\{#1\}}

\newcommand{\mat}[1]{\mathbf{#1}}

\newcommand{\fn}[1]{\mbox{\textsc{#1}}}
\newcommand{\rootp}{\sqrt{P}}
\newcommand{\basicattn}{\fn{Attention2D-NO}\xspace}
\newcommand{\basicattnbwd}{\fn{BwdAttention2D-NO}\xspace}
\newcommand{\attntext}{\fn{Attention2D}\xspace}
\newcommand{\attn}{\fn{Attention2D-O}\xspace}
\newcommand{\attnbwd}{\fn{BwdAttention2D-O}\xspace}
\newcommand{\flashattnfwd}{\fn{FlashAttn}\xspace}
\newcommand{\flashattnbwd}{\fn{BwdFlashAttn}\xspace}
\newcommand{\attnfix}{\fn{AttnFix}\xspace}
\newcommand{\gcs}{\fn{GthrCmptSctr}\xspace}
\newcommand{\gc}{\fn{GthrCmpt}\xspace}
\newcommand{\gcsbwd}{\fn{GthrCmptSctrBwd}\xspace}
\newcommand{\gcbwd}{\fn{GthrCmptBwd}\xspace}
\newcommand{\csbwd}{\fn{CmptSctrBwd}\xspace}

\newcommand{\qdim}{\fn{Q-dim}\xspace}
\newcommand{\kvdim}{\fn{KV-dim}\xspace}

\newcommand{\gex}[1]{\ge\!#1\times}

\hypersetup{
    pdftitle = {Attention2D},
    pdfauthor = {Venmugil Elango},
    colorlinks=true,
    linkcolor=cyan,
    citecolor=purple,
    urlcolor=violet,
}

\definecolor{colorp0}{rgb}{0,172,255}
\allowdisplaybreaks

\makeatletter
\DeclareRobustCommand\onedot{\futurelet\@let@token\@onedot}
\def\@onedot{\ifx\@let@token.\else.\null\fi\xspace}

\def\ie{\emph{i.e}\onedot}

\makeatother

\algnewcommand{\LeftComment}[1]{\Statex \(\triangleright\) #1}
\algnewcommand{\IfThen}[2]{
      \State \algorithmicif\ #1\ \algorithmicthen\ #2}

\title{\attntext: Communication Efficient Distributed Self-Attention Mechanism}
\author{Venmugil Elango \\ Microsoft}

\begin{document}

\maketitle

\begin{abstract}
Transformer-based models have emerged as a leading architecture for natural language processing,
natural language generation, and image generation tasks. A fundamental element of the
transformer architecture is self-attention, which allows the model to capture intricate
dependencies within the data. However, the self-attention mechanism also incurs significant
computational and memory costs, particularly for long sequences.

In this paper, we introduce \attntext, a novel approach that exploits parallelism along two
dimensions -- query and key/value -- of the self-attention operation. This method enables
efficient distribution and parallelization of computations across multiple devices. Our approach
facilitates \emph{asymptotically faster} training and inference phases compared to previous
methods, \emph{without relying on approximations or incurring additional computational or memory
overheads}. Furthermore, unlike existing techniques that struggle to scale with an increasing
number of processing units, our approach effectively scales with additional processing units.

Our experimental results confirm the effectiveness of our method in improving communication
efficiency and scalability. Compared to Ring
Attention~\citep{liu2023ringattentionblockwisetransformers}, our approach demonstrated up to
a $5\times$ performance boost on a GPT-3-like model using 64 NVIDIA A100 GPUs across 16 nodes,
and up to a $9.4\times$ performance boost on 64 NVIDIA H100 GPUs across 64 nodes.
\end{abstract}

\section{Introduction}
Transformer-based models~\citep{NIPS2017_3f5ee243} have shown tremendous success in showcasing
exceptional performance across wide range of AI applications and have emerged as the architecture of
choice in applications such as natural language processing and image classification. 
Long-context Transformers are essential for tackling a diverse array of AI challenges, including
processing books and high-resolution images, analyzing long videos and complex codebases, and
developing multi-agent AI solutions.
The use of vastly enlarged context has become a growing trend.
For example, GPT-4o~\citep{gpt4odocs2024} has a context length of 128,000 tokens, while Anthropic’s
Claude 3 model~\citep{claude3modelcard2024} features a context length of 200,000 tokens.

Self-attention operation is a key building block of Transformer models. However, self-attention is
slow and memory-intensive, especially for long contexts. 
This has motivated a surge of research on cutting down the memory and computation needs of
Transformer models.
Memory-efficient Transformers~\citep{dao2022flashattention, dao2023flashattention2} have emerged
that reduce the memory cost from quadratic to linear in terms of the context length, while
maintaining accuracy. However, little progress has been made in improving the communication cost.
Despite numerous efforts~\citep{MegatronLM, megatron_seqence_parallelism, li-etal-2023-sequence,
liu2023ringattentionblockwisetransformers, li2023lightseq, jacobs2023deepspeed, fang2024unified},
the communication cost remains linear with respect to context length, regardless of the
number of processing units used.
Specifically, for a Transformer model with $L$ layers, $M$ self-attention heads, and a head
dimension of $H$, these prior methods incur a training cost of $\bigo{BLNMH}$ per training step for
a batch size of $B$ and a context length of $N$.
The communication cost of these methods is independent of the number of processing units, and
therefore, they do not scale with an increasing number of processing units.
Table~\ref{tbl:costs} summarizes the communication costs associated with various parallelization
strategies for training.

\begin{table}
    {\centering
\caption{\label{tbl:costs}Communication costs of various parallelization strategies.
    Here, $B$ is batch size, $N$ is context length, $H$ is head dimension,
    $M$ is number of heads, and $P$ is number of processors.
    }
    \begin{tabular}{ll}
    \toprule
        \textbf{Parallelism} & \textbf{Communication cost per layer} \\
    \midrule
    Megatron-LM \citep{MegatronLM} & $\bigo{BNMH}$ \\
    Megatron-LM-SP \citep{megatron_seqence_parallelism} & $\bigo{BNMH}$ \\
    Sequence Parallelism \citep{li-etal-2023-sequence} & $\bigo{BNMH}$ \\
    Ring Attention \citep{liu2023ringattentionblockwisetransformers} & $\bigo{BNMH}$ \\
    Lightseq \citep{li2023lightseq} & $\bigo{BNMH}$ \\
    Ulysses$^\dagger$ \citep{jacobs2023deepspeed} & $\bigo{BNH}$ \\
    USP$^\ddag$ \citep{fang2024unified} & $\bigo{BNH}$ \\
        {\bf \attntext (ours)} & $\bigo{BNMH/\rootp}$ \\
    \bottomrule
\end{tabular}\\
    }
{\scriptsize $^\dagger$The parallelism degree of Ulysses cannot exceed the
  number of attention heads, $M$.\\
  $^\ddag$USP employs a combination of Ring Attention and Ulysses parallelism.
  Consequently, the parallelism degree assigned to the Ulysses dimension cannot
  surpass the number of attention heads, $M$.
    }
\end{table}

In this paper, we propose \attntext that achieves sub-linear communication cost, which decreases as
the number of processing units increases. This leads to asymptotically lower communication costs and
better scalability compared to the current state of the art.
As seen in Table~\ref{tbl:costs}, the communication cost of earlier methods remains unchanged
regardless of the number of devices. In contrast, communication cost in \attntext (ours) decreases in
proportion to the square root of the number of devices used.
We are able to achieve this without any approximation by
leveraging the associativity property of softmax statistics computation and parallelizing the
self-attention computation across both query tokens and key-value token pairs.
In contrast, various prior works~\citep{li-etal-2023-sequence,
liu2023ringattentionblockwisetransformers, li2023lightseq} only parallelize the self-attention
computation across various query tokens\footnote{Although some of these techniques distribute data
across both query and key-value pairs, the computation is parallelized solely across the queries.
Key and value tokens are entirely all-gathered during the computation. This leads to
memory savings, but not communication cost savings. Refer Subsection~\ref{sec:1dpar} for more
details.}, resulting in sub-optimal communication costs.

Our approach is applicable to both the training and inference stages. The inference phase of
decoder-only transformer models~\citep{gpt3} consist of two distinct computational phases: the
prompt phase (also known as the prefill phase) and the token generation phase (also known as the
decode phase). During the prompt phase, all $N$ tokens from the input prompt are processed in
parallel through the model to generate the first output token, making this phase computationally
intensive. After the prompt computation, the decode phase begins in an auto-regressive manner, where
each subsequent token of a single request is generated one at a time based on the forward pass of
the preceding token. Consequently, the query length in the prompt phase equals the context size $N$,
while in the decode phase, the query length is typically $1$. Similar to previous
methods~\citep{li-etal-2023-sequence, liu2023ringattentionblockwisetransformers, li2023lightseq},
our approach is mainly suitable for the prompt phase of inference in decoder-only
transformer models, where the query length matches the context size.

\section{\label{sec:background}Background}
In this section, we provide the necessary background on self-attention and the one-dimensional
parallelization scheme for distributed self-attention computation.

\paragraph{Notation.}
We use boldface uppercase letters to denote matrices, standard uppercase letters for vectors, and
lowercase letters for scalars. Given a matrix $\mat{A}$,
$a_{i,j}$ refers to the element of $\mat{A}$ in row $i$ and column $j$.
Our indexing is 0-based. Thus, the first element of the matrix $\mat{A}$ is denoted as $a_{0,0}$.
Additionally, we utilize standard uppercase letters to represent various model and computation
parameters. These parameters are summarized in Table~\ref{tbl:params}.

\begin{table}
    \centering
    \caption{\label{tbl:params}Notation for parameters.}
    \begin{tabular}{lll}
        \toprule
        $B$: Batch size & $N$: Context length & $L$: Number of layers \\
        $H$: Head dimension & $M$: Number of heads & $P$: Number of devices \\
        \bottomrule
    \end{tabular}
\end{table}

\subsection{Standard self-attention}
A standard self-attention operation\footnote{For the sake of clarity, we omit the scaling of
$\mat{Q}\mat{K}^T$ (typically by $1/\sqrt{H}$) and dropout applied to the softmax output as this
does not affect the ideas presented.}~\citep{NIPS2017_3f5ee243} for a single head is given by the
following equation:
\begin{equation}
    \mat{S} = \mat{Q}\mat{K}^T + \mat{X}; \quad 
    \mat{O} = \fn{Softmax}(\mat{S})\mat{V}
    \label{eq:std_attn}
\end{equation}
where, $\mat{Q}, \mat{K},
\mat{V} \in \rspace{N\times H}$ are the query, key and value matrices, each containing $N$ tokens,
where each token is of size $H$. The matrices $\mat{S} \in \rspace{N\times N}$ and $\mat{O} \in
\rspace{N\times H}$ represent the attention weights and output, respectively. The $\fn{Softmax}$
operation is a numerically stable variant of the softmax function. Additionally, $\mat{X} \in
\rspace{N\times N}$ is the mask matrix that determines which tokens in a sequence can be
attended to during the computation of attention scores.
For causal language models such as GPT, this mask is given by:
\begin{equation*}
    x_{i,j} = 
    \begin{cases}
        0 & \text{if } i \geq j \\
        -\infty & \text{if } i < j
    \end{cases}
\end{equation*}
For masked language models such as BERT, $x_{i,j}$ is set to $-\infty$ at randomly selected masked
locations and $0$ at other locations during training.
Eq.~(\ref{eq:std_attn}) can be written in element form as follows:
\begin{equation*}
    s_{i,r} = \sum_{h=0}^{H-1} q_{i,h}k_{r,h} + x_{i,r},
    \quad m_i = \max_{r=0}^{N-1} s_{i,r},
    \quad d_i = \sum_{r=0}^{N-1} e^{s_{i,r}-m_i},
\end{equation*}
\begin{equation}
    o_{i,j}  = 
             \sum_{r=0}^{N-1} \frac{e^{s_{i,r}-m_i}}{d_i}v_{r,j}
             = \frac{1}{d_i} \sum_{r=0}^{N-1} e^{s_{i,r}-m_i}v_{r,j}
            \label{eq:o_ik_orig}
\end{equation}

\subsection{\label{sec:flashattn}Memory-efficient self-attention}
A standard self-attention operation requires $O(N^2)$ memory per batch and per head to store the
intermediate matrix $\mat{S}$.
Previous studies ~\citep{online_softmax, memory_efficient_self_attn, dao2022flashattention,
dao2023flashattention2} demonstrated that this operation can be performed with constant memory,
without materializing the full $\mat{S}$ matrix and without any approximations, as follows:
Let $o'_{i,j,B}$ denote the $j$-th element in the self-attention output for the query at index $i$ with the first $B$ key
and value tokens, then, 
\begin{equation*}
    m'_{i,B} = \max_{r=0}^{B-1} s_{i,r}, \quad
    n'_{i,j,B} = \sum_{r=0}^{B-1} e^{s_{i,r}-m'_{i,B}} v_{r,j}, \quad
    d'_{i,B} = \sum_{r=0}^{B-1} e^{s_{i,r}-m'_{i,B}}, \quad
    o'_{i,j,B} = \frac{n'_{i,j,B}}{d'_{i,B}}
\end{equation*}
Then, by utilizing the associative properties of $\max$ and addition, the self-attention output for the
query index $i$ with the first $2B$ key and value tokens, $o'_{i,j,2B} = n'_{i,j,2B} / d'_{i,2B}$
can be obtained from $o'_{i,j,B}$ (without the intermediates $s_{i,0},\dots,s_{i,B-1}$) as follows:
\begin{equation*}
    \begin{gathered}
    m'_{i,2B} = \max_{r=0}^{2B-1} s_{i,r} 
        = \max(m'_{i,B}, \max_{r=B}^{2B-1} s_{i,r}) \\
    n'_{i,j,2B} = \sum_{r=0}^{2B-1} e^{s_{i,r}-m'_{i,2B}} v_{r,j} 
        = e^{m'_{i,B}-m'_{i,2B}} n'_{i,j,B} + \sum_{r=B}^{2B-1} e^{s_{i,r}-m'_{i,2B}} v_{r,j} \\
    d'_{i,2B} = \sum_{r=0}^{2B-1} e^{s_{i,r}-m'_{i,2B}} 
        = e^{m'_{i,B}-m'_{i,2B}} d'_{i,B} + \sum_{r=B}^{2B-1} e^{s_{i,r}-m'_{i,2B}} \\
    \end{gathered}
\end{equation*}
The complete self-attention output $o_{i,j}$ can thus be generated in $N/B$ steps with constant
memory.

\subsection{\label{sec:1dpar}1D parallel self-attention}
\begin{figure}
    \centering
    \includegraphics[width=0.5\textwidth]{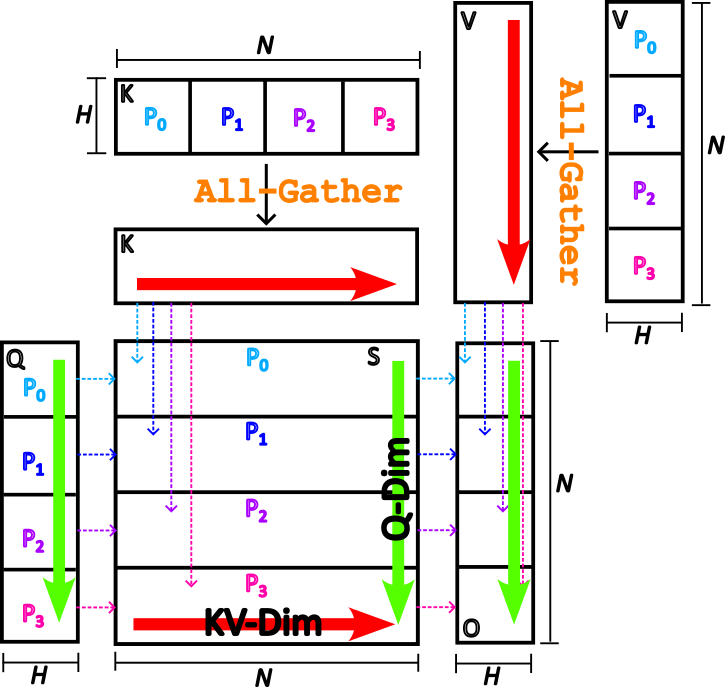}
    \caption{\label{fig:attn1d}1D parallel self-attention computation distributed across four
    processors. For simplicity, masking and softmax operations are omitted. 
    Initially, $\mat{Q}$, $\mat{K}$, and $\mat{V}$ are evenly distributed among all processors.
    The two computation dimensions, \qdim and \kvdim, are indicated with green and red arrows,
    respectively.
    Computation is fully parallel along the \qdim, while the \kvdim is sequential due to the data
    dependence enforced by the row-wise softmax operation. Dotted arrows indicate the inputs
    required by different processors to compute their blocks of $\mat{S}$ and $\mat{O}$. All
    processors must collectively perform an all-gather operation on $\mat{K}$ and $\mat{V}$ matrices
    to acquire the necessary input data for their computation.
    The communication cost of this operation is $\bigo{NH}$ per batch and per head.}
\end{figure}
As seen in Eq.~(\ref{eq:o_ik_orig}), computation of self-attention for different query tokens are
independent of each other. The self-attention computation can thus be parallelized by assigning the
self-attention output computation for different queries to different processors.
This is illustrated in Figure~\ref{fig:attn1d} (with masking and softmax operations omitted for
clarity) using four processors. 
The computation occurs in a two-dimensional iteration space (\ie the computation is composed of two
nested loops), denoted by \qdim dimension (green arrows) and \kvdim dimension (red arrows) in
Figure~\ref{fig:attn1d}.  Of these two dimensions, the \qdim dimension is readily parallelizable and
can be distributed across multiple processors.
In contrast, the \kvdim dimension, due to data dependencies from the row-wise softmax operation, is
sequential, preventing parallel computation along \kvdim, resulting in one-dimensional parallelism.
Although memory-efficient self-attention techniques eliminate the need for quadratic memory, the
sequential dependency along the \kvdim remains, since the computation of any block $o'_{i,j,kB}$
depends on the values computed for $o'_{i,j,(k-1)B}$.
In Figure~\ref{fig:attn1d}, different processors compute different blocks of the
output matrices $\mat{S}$ and $\mat{O}$, split along the \qdim. Each processor requires the full
$\mat{K}$ and $\mat{V}$ matrices, which are typically evenly distributed among all processors due to
memory constraints. This necessitates a collective all-gather operation on $\mat{K}$ and $\mat{V}$
matrices, resulting in a communication cost of $\bigo{NH}$ per batch and per head, linear in the context
length and independent of the number of processors.  Previous techniques such as
Ring Attention~\citep{li-etal-2023-sequence, liu2023ringattentionblockwisetransformers} attempt to
mitigate this overhead to a certain extent by overlapping the computation and communication.
However, the need to collectively communicate the full $\mat{K}$ and $\mat{V}$ matrices persists,
resulting in a suboptimal communication cost. As language model context lengths increase, this
becomes a fundamental bottleneck in distributed self-attention operation, which is unavoidable with
parallelism along just a single dimension.

\section{2D Parallel Self-attention}
As outlined in Section~\ref{sec:background}, the self-attention computation is inherently sequential
along the \kvdim.
In this section, we describe how Eq.~(\ref{eq:o_ik_orig}) can be rewritten to expose parallelism
along both the \qdim and \kvdim dimensions. This allows us to efficiently parallelize the
self-attention computation across both dimensions, resulting in \emph{asymptotically lower
communication costs without approximations}. 
We first introduce a non-overlapping version of our method, \basicattn, in
Subsection~\ref{sec:basic_attn}, where computation and communication are performed in separate,
non-overlapping phases.
In Subsection~\ref{sec:cost_analysis}, we derive the communication and memory costs of our method.
In Subsection~\ref{sec:tiled_attn}, we further develop \basicattn into \attn, which overlaps
computation and communication to achieve additional practical benefits in certain scenarios.

\subsection{\label{sec:parallel_kv}Exposing parallelism along \kvdim}
In order to expose parallelism along \kvdim, by utilizing the associative and commutative properties
of $\max$ and addition operations, we rewrite Eq.~(\ref{eq:o_ik_orig}) as follows:
Let $R_1,R_2,\dots$ be a disjoint partition of $\set{0,1,\dots,N-1}$, \ie,
${R_1 \cup R_2 \cup \dots}=\set{0,1,\dots,N-1}$ and $R_x \not= R_y \implies R_x
\cap R_y=\emptyset$. Then, the self-attention output of a query at index $i$ with the set of key and
value indices in $R_x$, $o'_{i,j,R_x} = n'_{i,j,R_x}/d'_{i,R_x}$, where,
\begin{equation}
    m'_{i,R_x} = \max_{r \in R_x} s_{i,r}, \quad
    n'_{i,j,R_x} = \sum_{r \in R_x} e^{s_{i,r}-m'_{i,R_x}} v_{r,j}, \quad
    d'_{i,R_x} = \sum_{r \in R_x} e^{s_{i,r}-m'_{i,R_x}}
    \label{eq:o_ik_rx}
\end{equation}
Given two partial self-attention outputs $o'_{i,j,R_x}$ and $o'_{i,j,R_y}$, their combined output
$o'_{i,j,R} = n'_{i,j,R}/d'_{i,R}$, where $R = R_x \cup R_y$, can be computed as follows:
\begin{align*}
        m'_{i,R} &= \max_{r \in R} s_{i,r} 
                   = \max(\max_{r \in R_x} s_{i,r}, \max_{r \in R_y} s_{i,r})
                   = \max(m'_{i,R_x}, m'_{i,R_y}) \\
        n'_{i,j,R} &= \sum_{r \in R} e^{s_{i,r}-m'_{i,R}} v_{r,j} \\
                    &= \sum_{r \in R_x} e^{s_{i,r}-m'_{i,R}} v_{r,j} 
                        + \sum_{r \in R_y} e^{s_{i,r}-m'_{i,R}} v_{r,j} \\
                    &= e^{m'_{i,R_x} - m'_{i,R}} \sum_{r \in R_x} e^{s_{i,r}-m'_{i,R_x}} v_{r,j}
                        + e^{m'_{i,R_y} - m'_{i,R}} \sum_{r \in R_y} e^{s_{i,r}-m'_{i,R_y}} v_{r,j} \\
                    &= e^{m'_{i,R_x} - m'_{i,R}} n'_{i,j,R_x} 
                        + e^{m'_{i,R_y} - m'_{i,R}} n'_{i,j,R_y} \\
        d'_{i,R} &= \sum_{r \in R} e^{s_{i,r}-m'_{i,R}} \\
                  &= \sum_{r \in R_x} e^{s_{i,r}-m'_{i,R}} + \sum_{r \in R_y} e^{s_{i,r}-m'_{i,R}} \\
                  &= e^{m'_{i,R_x} - m'_{i,R}} \sum_{r \in R_x} e^{s_{i,r}-m'_{i,R_x}} 
                        + e^{m'_{i,R_y} - m'_{i,R}} \sum_{r \in R_y} e^{s_{i,r}-m'_{i,R_y}} \\
                  &= e^{m'_{i,R_x} - m'_{i,R}} d'_{i,R_x} + e^{m'_{i,R_y} - m'_{i,R}} d'_{i,R_y}
\end{align*}
This enables us to parallelize \kvdim along with \qdim by distributing the computation of different
partial outputs $o'_{i,j,R_k}$ to different devices. These partial outputs are then combined
through a reduction operation using the above set of equations to achieve the complete self-attention
output.

\subsection{\label{sec:basic_attn}2D Parallel Self-attention Computation Without Overlapping
Communication}
\begin{figure}
    \centering
    \includegraphics[width=0.8\textwidth]{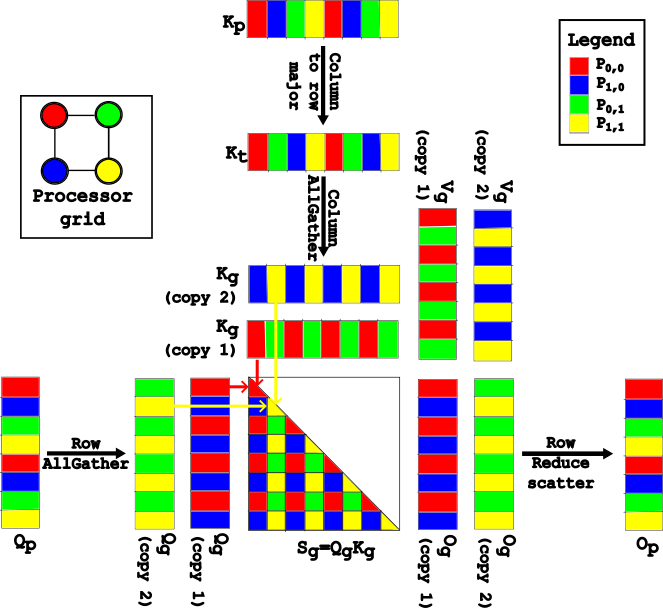}
    \caption{\label{fig:attn2d_basic} This figure illustrates the non-overlapping 2D parallel
    self-attention computation using four devices and eight tokens. The four devices are arranged in
    a $2 \times 2$ mesh, as depicted in the top left corner. Initially, the $\mat{Q}$, $\mat{K}$,
    and $\mat{V}$ matrices are cyclically distributed in a 1D column-major format. The $\mat{K}$ and
    $\mat{V}$ matrices are then redistributed from a column-major to a row-major format.
    Subsequently, the $\mat{Q}$ matrix is all-gathered within the subset of processors in the same
    row, while the $\mat{K}$ and $\mat{V}$ matrices are all-gathered within the subset of processors
    in the same column. Each processor then performs the flash-attention operation using its local
    copies of $\mat{Q}$, $\mat{K}$, and $\mat{V}$. For clarity, the masking and softmax operations
    are not shown. As indicated by different colors in the figure, the computation of the attention
    score $s_{i,j}$ is performed by the unique processor that holds the input tokens $q_i$ and
    $k_j$. For instance, the score $s_{0,0}$ is computed by processor
    \textcolor{red}{$p_{0,0}$}, which is the only processor containing both
    $q_0$ and $k_0$ (illustrated with \textcolor{red}{$\rightarrow$} and
    \textcolor{red}{$\downarrow$} arrows). The
    computation is load balanced due to cyclic distribution, with each processor computing exactly
    ten elements of $\mat{S}$.  Finally, the partial outputs are reduce-scattered among the subset
    of processors within the same row, resulting in the complete output with tokens distributed in
    a 1D fashion similar to the inputs.
    }
\end{figure}

Building on Subsection~\ref{sec:parallel_kv}, we now detail the forward pass of the self-attention
operation utilizing 2D parallelism. The pseudocode for this procedure is outlined in
Alg.~\ref{alg:uattn_fwd}. Figure~\ref{fig:attn2d_basic} illustrates this process with an example
involving 8 tokens and 4 GPUs.
Given $P$ processors, we conceptualize the processors as forming a $\rootp \times \rootp$
square grid, where $p_{r,c}$ corresponds to the processor in row $r$ and column $c$ of the grid.
While a square grid is not strictly necessary for our method to function, it generally maximizes the
reduction in communication volume since the communication volume along both rows and columns is
typically the same. However, if the communication
volume varies across dimensions, such as when the $\mat{KV}$ matrix is quantized
differently from the $\mat{Q}$ matrix, resulting in communication imbalance, a rectangular grid
would be more optimal. Additionally, if $P$ is not a perfect square, the processors can be arranged in
a rectangular grid that approximates the optimal size to minimize communication volume.

\begin{algorithm}
\begin{algorithmic}[1]
    \Require 
    $\mat{Q_p}, \mat{K_p}, \mat{V_p} \in \mathbb{R}^{(N/P)\times H}$: Query, key, and value matrices
    cyclically distributed;\\
    $p_{r,c}$: Current processor; $P$: Total number of processors

    \State $\mat{K_t}, \mat{V_t} \gets$ \fn{SendRecv}($\set{\mat{K_p}, \mat{V_p}}, p_{c,r}, p_{c,r}$)
    \Comment{Column $\rightarrow$ row-major distribution} \label{alg:uattn_fwd:send_kv}

    \State $\mat{Q_g}\gets$ \fn{AllGather}($\mat{Q_p}, \set{p_{r,i}~|~i\in[0,\rootp)}$)
    \Comment{Row all-gather}\label{alg:uattn_fwd:gather_q}
    \State $\mat{K_g}, \mat{V_g} \gets$ \fn{AllGather}($\set{\mat{K_t}, \mat{V_t}},
    \set{p_{i,c}~|~i\in[0,\rootp)}$) \Comment{Column all-gather}\label{alg:uattn_fwd:gather_kv}

    \State $M_g, \mat{N_g}, D_g \gets \flashattnfwd(\mat{Q_g},
    \mat{K_g}, \mat{V_g}$) \Comment{See Section~\ref{sec:flashattn} \&
    \citep{dao2023flashattention2}} \label{alg:uattn_fwd:local_attn}

    \State $M_p, \mat{N_p}, D_p \gets $\fn{RowReduceScatter}($M_g$, $\mat{N_g}$, $D_g$, $p_{r,c}$,
    $P$) \Comment{Reduce-scatter using Alg.~\ref{alg:reduce_scatter}}
    \label{alg:uattn_fwd:reduce}

    \State $\mat{O_p} \gets \fn{Diag}(D_p^{-1})\mat{N_p}$
    \label{alg:uattn_fwd:O_p}
    \State \fn{SaveForBackprop}($\mat{Q_p},\mat{K_t},\mat{V_t},\mat{O_p},M_p,D_p$)
    \State \Return $\mat{O_p}$
\end{algorithmic}
    \caption{\label{alg:uattn_fwd}\basicattn: Forward pass for non-overlapping 2D parallel
    self-attention.}
\end{algorithm}

The inputs to the self-attention operator, matrices $\mat{Q}, \mat{K}, \mat{V} \in \rspace{N \times
H}$, are initially distributed evenly and cyclically among all processors in a column-major 1D
manner when they arrive from the previous layer, \ie, for a processor $p_{r,c}$, its matrices
$\mat{Q_p}, \mat{K_p}, \mat{V_p} \in \rspace{(N/P) \times H}$ contain query, key, and value tokens,
respectively, for indices $Q_p(r,c) = \set{r + \rootp c + iP~|~i \in [0,N/P)}$.
The distribution of $\mat{K}$ and $\mat{V}$ matrices are first transposed from column-major to
row-major form in line~\ref{alg:uattn_fwd:send_kv}, \ie, for a processor $p_{r,c}$, $\mat{K_t}$ and
$\mat{V_t}$ contain the keys and values, respectively, for token indices $K_t(r,c) = \set{\rootp
r + c + iP~|~i \in [0,N/P)}$.
In line~\ref{alg:uattn_fwd:gather_q}, the matrix $\mat{Q_p}$ is collectively gathered among the
subset of processors $\set{p_{r,i}~|~i\in[0,\rootp)}$ that are within the same row of the processor grid. Consequently, $\mat{Q_g}
\in\rspace{(N/\rootp) \times H}$ of a processor $p_{r,c}$ contains the query tokens for indices
$Q_g(r) = \cup_{j=0}^{\rootp-1} Q_p(r,j) = \set{r + i\rootp~|~i \in [0,N/\rootp)}$.  Similarly, in
line~\ref{alg:uattn_fwd:gather_kv}, the matrices $\mat{K_t}$ and $\mat{V_t}$ are collectively
gathered among subset of processors within the same columns of the processor grid, resulting in
token indices $K_g(c) = \cup_{j=0}^{\rootp-1} K_t(j,c) = \set{c + i\rootp~|~i \in [0,N/\rootp)}$.

By the end of line~\ref{alg:uattn_fwd:gather_kv}, a processor $p_{r,c}$ contains all the necessary
query, key, and value tokens required to compute the partial values $m'_{i,R_x}$, $n'_{i,j,R_x}$,
and $d'_{i,R_x}$ of Eq.~(\ref{eq:o_ik_rx}) for $i \in Q_g(r)$, $j \in [0,H)$, and $R_x=K_g(c)$.  The
function \flashattnfwd in line~\ref{alg:uattn_fwd:local_attn} uses the flash-attention technique
developed by~\citet{dao2023flashattention2} to perform this computation in a memory efficient manner
as described in Section~\ref{sec:flashattn}. For simplicity, the masking function is omitted from the
pseudocode. As illustrated in Figure~\ref{fig:attn2d_basic}, when causal masking is applied, the
mask for each processor, similar to the original mask, also forms a lower triangular matrix because of
the cyclic distribution~\footnote{To accommodate the off-diagonal causal masks present in our
approach, we made minor modifications to the original flash-attention kernel
(https://github.com/Dao-AILab/flash-attention/tree/v2.4.2).}.
Upon computing the partial outputs across various processors, it is necessary to aggregate these
outputs as shown in Subsection~\ref{sec:parallel_kv} to derive the complete self-attention output. The partial
outputs corresponding to a single query are distributed among different processors within the same
row of the processor grid. Consequently, the reduction process is confined to a subset of processors
sharing the same row. Furthermore, the outputs should be cyclically distributed among all available
processors in a one-dimensional (1D) manner, similar to the input distribution. This is accomplished
through a standard reduce-scatter collective operation, using \attnfix (Alg.~\ref{alg:reduce_op}) as
the reduction operator and cyclic distribution as the distribution pattern.
For completeness, Appendix~\ref{sec:row_reduce_scatter} includes pseudocode for the standard
ring-based reduce-scatter operation, tailored for cyclic data distribution within the rows of
a two-dimensional processor grid.
Finally, in line~\ref{alg:uattn_fwd:O_p}, $\mat{N_p}$ is divided by $D_p$ to obtain the output
$\mat{O_p} \in \rspace{(N/P) \times H}$.\footnote{In our actual implementation, we work with the
log-sum-exponent $L_p = M_p + \log(D_p)$ instead of $M_p$ and $D_p$. To keep the description simple,
we omit these details in the discussion. For more information, see~\citep[Section
3.1.1]{dao2023flashattention2}.}
Additionally, the vectors $M_p$ and $D_p$, along with $\mat{O_p}, \mat{Q_p}, \mat{K_t}, \mat{V_t}
\in \rspace{(N/P) \times H}$ are saved for gradient computation during backward pass.

\begin{algorithm}
\begin{algorithmic}[1]
    \Require $M_1, \mat{N_1}, D_1$: First set of partial self-attention outputs;\\
             $M_2, \mat{N_2}, D_2$: Second set of partial self-attention outputs
    \State $M \gets \max(M_1, M_2)$\Comment{Elementwise max}\label{alg:reduce_op:M}
    \State $E_1 \gets \exp(M_1-M)$, $E_2 \gets \exp(M_2-M)$ \Comment{Elementwise exponents}
    \label{alg:reduce_op:E}
    \State $\mat{N} \gets$ \fn{Diag}($E_1$)$\mat{N_1}$
    + \fn{Diag}($E_2$)$\mat{N_2}$ \label{alg:reduce_op:N}
    \State $D \gets E_1 \odot D_1 + E_2 \odot D_2$ \Comment{$\odot$: Elementwise multiplication}
    \label{alg:reduce_op:D}
    \State \Return $M, \mat{N}, D$
\end{algorithmic}
    \caption{\label{alg:reduce_op}\attnfix: Combines two partial self-attention outputs as
    shown in Subsection~\ref{sec:parallel_kv}.}
\end{algorithm}

The pseudocode for backward pass is detailed in Alg.~\ref{alg:uattn_bwd}. Similar to the forward
pass, the backward pass is executed in a 2D parallel manner.  Since the intermediate attention
weights and scores are not stored during the forward pass, they need to be recalculated during the
backward pass.
To facilitate this, the query matrix is gathered within each processor grid row in
line~\ref{alg:uattn_bwd:gather_q}, and the key and value matrices are gathered within each processor
grid column in line~\ref{alg:uattn_bwd:gather_kv}.
Additionally, the output $\mat{O_p}$ and its gradient $\mat{dO_p}$ are also gathered within each
processor grid row.  The function \flashattnbwd utilizes the algorithm by~\citet[Section
3.1.2]{dao2023flashattention2} to recompute the attention weights and scores and to calculate the
gradients of the query, key, and value matrices in a memory-efficient fashion.  The query gradient
matrix is then reduce-scattered, using summation as the reduction operation, within the rows (in
line~\ref{alg:uattn_bwd:reduce_q}), and the key and value matrices are reduce-scattered within the
columns (in line~\ref{alg:uattn_bwd:reduce_kv}) of the processor grid.  Finally, the key and value
gradients are transposed from row-major to column-major distribution in
line~\ref{alg:uattn_bwd:send_kv}.

\begin{algorithm}
\begin{algorithmic}[1]
    \Require
    $\mat{Q_p}, \mat{O_p}, \mat{dO_p} \in \mathbb{R}^{(N/P) \times H}$: Query, output, and gradient
    with column-major distribution;\\
    $\mat{K_t}, \mat{V_t} \in \mathbb{R}^{(N/P) \times H}$: Key and value matrices with row-major
    distribution;\\
    $M_p, D_p \in \rspace{N/P}$: Softmax statistics with column-major distribution;\\
    $p_{r,c}$: Current processor; $P$: Total number of processors

    \State $\mat{Q_g}, \mat{O_g}, \mat{dO_g}, M_g, D_g \gets$ \fn{AllGather}($\set{\mat{Q_p}, \mat{O_p},
    \mat{dO_p}, M_p, D_p}, \set{p_{r,i}~|~i\in[0,\rootp)}$)
    \Comment{Row all-gather}\label{alg:uattn_bwd:gather_q}
    \State $\mat{K_g}, \mat{V_g} \gets$ \fn{AllGather}($\set{\mat{K_t}, \mat{V_t}},
    \set{p_{i,c}~|~i\in[0,\rootp)}$) \Comment{Column all-gather}\label{alg:uattn_bwd:gather_kv}

    \State $\mat{dQ_g}, \mat{dK_g}, \mat{dV_g} \gets \flashattnbwd(\mat{Q_g}$,
    $\mat{K_g}$, $\mat{V_g}$, $\mat{O_g}$, $\mat{dO_g}$, $M_g$, $D_g$) 
    \Comment{Refer \citep{dao2023flashattention2}} \label{alg:uattn_bwd:local_attn}

    \State $\mat{dQ_p} \gets $\fn{ReduceScatter}($\mat{Q_g}$, $\set{p_{r,i}~|~i\in[0,\rootp)}$)
    \Comment{Row reduce-scatter} \label{alg:uattn_bwd:reduce_q}
    \State $\mat{dK_t}, \mat{dV_t} \gets $\fn{ReduceScatter}($\set{\mat{K_g}, \mat{V_g}}$,
    $\set{p_{i,c}~|~i\in[0,\rootp)}$) \Comment{Column reduce-scatter}
    \label{alg:uattn_bwd:reduce_kv}

    \State $\mat{dK_p}, \mat{dV_p} \gets$ \fn{SendRecv}($\set{\mat{dK_t}, \mat{dV_t}}, p_{c,r}, p_{c,r}$)
    \Comment{Row $\rightarrow$ column-major distribution} \label{alg:uattn_bwd:send_kv}

    \State \Return $\mat{dQ_p}, \mat{dK_p}, \mat{dV_p}$
\end{algorithmic}
    \caption{\label{alg:uattn_bwd}\basicattnbwd: Backward pass for \basicattn.}
\end{algorithm}

\subsection{\label{sec:cost_analysis}Communication and Memory Cost Analysis}
\paragraph{Communication cost.}
The data loader feeds the input tokens in a cyclic 1D distribution along the sequence length
dimension, eliminating the need for any initial distribution of the inputs.
Linear layers, such as the QKV-projection and the feed-forward network (FFN), are
parallelized using ZeRO-3~\citep{zero} / FSDP~\citep{fsdp} style data parallelism with both the
batch and sequence length dimensions collectively considered for data parallelism.
Since this is an activation-stationary parallelism scheme, the communication involved includes
all-gather during the forward pass and reduce-scatter during backward pass for the weight matrices.
Consequently, for a transformer model with $L$ layers, the overall communication cost incurred by
the linear layers is $\bigo{LM^2H^2}$.
The communication cost for the self-attention operation, considering a single batch, single layer,
and single head, is as follows:
The initial redistribution of $\mat{K_p}$ and $\mat{V_p}$ from column-major to row-major format
(line~\ref{alg:uattn_fwd:send_kv} in Alg.~\ref{alg:uattn_fwd}) requires a simple point-to-point
communication, incurring a cost of $\bigo{NH/P}$.
The all-gather operations in lines~\ref{alg:uattn_fwd:gather_q} and
\ref{alg:uattn_fwd:gather_kv} involve the collective communication of $\bigo{NH/P}$ words of data by
each processor among $\rootp$ processors, 
incurring a cost of $\bigo{NH/\rootp}$. Similarly, the communication
cost of reduce-scatter operation in
line~\ref{alg:uattn_fwd:reduce} is $\bigo{NH/\rootp}$. 
During the backward pass, the collective communication operations in
lines~\ref{alg:uattn_bwd:gather_q}, \ref{alg:uattn_bwd:gather_kv}, \ref{alg:uattn_bwd:reduce_q}
and~\ref{alg:uattn_bwd:reduce_kv}
in Alg.~\ref{alg:uattn_bwd} each incur a cost of $\bigo{NH/\rootp}$, while the point-to-point
communication cost of line~\ref{alg:uattn_bwd:send_kv} is $\bigo{NH/P}$.
Therefore, the overall communication cost of the 2D parallel self-attention operation for a single batch,
single layer, and single head is $\bigo{NH/\rootp}$.
For a model with $L$ layers, $M$ heads, and a batch size of $B$, the total communication cost of
the self-attention operation for a single training step is $\bigo{BLMNH/\rootp}$.
Consequently, the combined communication cost for both self-attention and non-self-attention
operations for a single training step is $\bigo{LM^2H^2 + BLMNH/\rootp}$ which reduces to
$\bigo{BLMNH/\rootp}$ when $N > MH$ for long sequence length training.
In contrast, as outlined in Table~\ref{tbl:costs}, existing approaches such as
Megatron-LM~\citep{MegatronLM, megatron_seqence_parallelism} and Ring
Attention~\citep{li-etal-2023-sequence, liu2023ringattentionblockwisetransformers} incur
a communication cost of $\bigo{BLMNH}$ that increases linearly with the sequence length $N$,
irrespective of the number of processors used.

\paragraph{Memory cost.}
The matrices $\mat{Q_p}$, $\mat{K_p}$, $\mat{V_p}$, $\mat{K_t}$, and $\mat{V_t}$ used in the forward
pass, along with $\mat{dQ_p}$, $\mat{dK_p}$, $\mat{dV_p}$, $\mat{dK_t}$, $\mat{dV_t}$, and
$\mat{dO_p}$ used in the backward pass, each require memory of size $\bigo{NH/P}$.  The intermediate
gathered matrices $\mat{Q_g}$, $\mat{K_g}$, $\mat{V_g}$, and $\mat{N_g}$ in the forward pass, and
$\mat{dQ_g}$, $\mat{dK_g}$, $\mat{dV_g}$, and $\mat{dO_g}$ in the backward pass, each require memory
of size $\bigo{NH/\rootp}$. However, these intermediate matrices are freed at the end of each layer's
computation and are not stored throughout the entire training step.  The vectors $M_g$ and $D_g$
each require a memory of $\bigo{N/\rootp}$, while the vectors $M_p$ and $D_p$ each require memory of
size $\bigo{N/P}$.  Among these, only the matrices $\mat{Q_p}$, $\mat{K_p}$, $\mat{V_p}$,
$\mat{O_p}$ and the vectors $M_p$ and $D_p$ of each layer are stored for the backward pass and
persist throughout the entire training step, while the rest are freed at the end of each layer.
Therefore, for a single training step with a batch size of $B$, $L$ layers, and $M$ attention heads,
the total memory cost is $\bigo{BLMNH/P + BNH/\rootp}$. When $LM > \rootp$, this simplifies to
$\bigo{BLMNH/P}$, which is the same as the existing approaches such as Megatron-LM and Ring
Attention.
For the non-self-attention layers, such as the QKV-projection and the FFN, the activation
sizes become more significant than the weights at long sequence lengths, resulting in a total memory
cost of $\bigo{BLMNH/P}$ when using ZeRO-3/FSDP style parallelism.
Consequently, the total memory cost incurred by our approach is $\bigo{BLMNH/P}$, which is same as
the existing techniques.

\subsection{\label{sec:tiled_attn}2D Parallel Self-attention Computation With Overlapping
Communication}
\begin{figure}
    \centering
    \includegraphics[width=\textwidth]{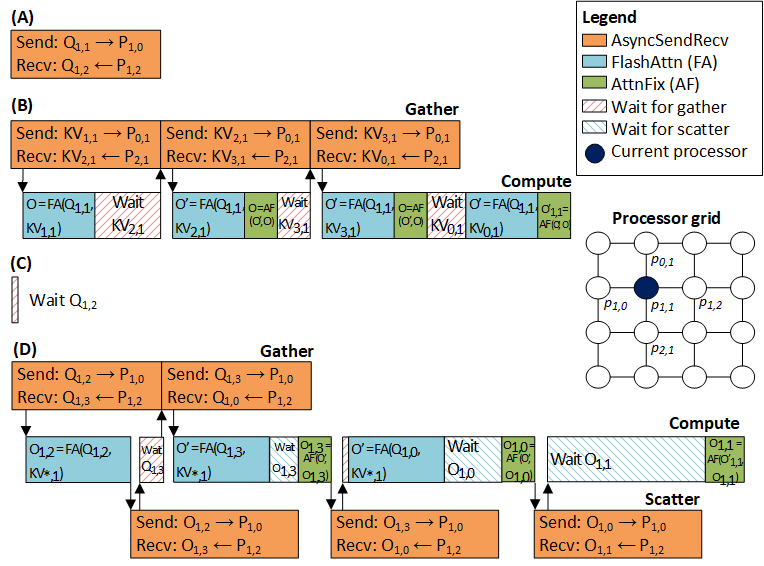}
    \caption{\label{fig:attn2d}
    This figure demonstrates the 2D parallel self-attention computation with overlapping
    communication using 16 devices arranged in a $4 \times 4$ mesh. The process, focused on device
    $p_{1,1}$, involves four steps:
    (A) Device $p_{1,1}$ asynchronously sends its local block $\mat{Q_{1,1}}$ to its left neighbor
    and receives block $\mat{Q_{1,2}}$ from its right neighbor.
    (B) Overlapping gather+compute (\gc) is performed in a ring fashion within the processor grid column. During
    each iteration $j\in [1,\rootp)$, a key-value block is sent to the top neighbor and
    received from the bottom neighbor, and $\flashattnfwd(\mat{Q_{1,1}}, \mat{KV_{j\% \rootp,1}})$ is
    executed. After $\rootp-1=3$ steps, all key-value blocks $\mat{KV_{*,1}}$
    are available at $p_{1,1}$.
    The output $\mat{O'_{1,1}}$ contains the partial result of computing self-attention with
    $\mat{Q_{1,1}}$ and $\mat{KV_{*,1}}$.
    (C) The processor waits until $\mat{Q_{1,2}}$ is fully received from $p_{1,2}$.
    (D) Overlapping gather+compute+scatter (\gcs) is performed in a ring fashion within the processor grid row.
    During each iteration $i \in [2,\rootp)$, a query block is sent to the left neighbor and
    received from the right neighbor, and $\flashattnfwd(\mat{Q_{1,i\% \rootp}}, \mat{KV_{*,1}})$ is
    executed. At the end, $p_{1,1}$ sends partial result $\mat{O_{1,0}}$ to the
    left neighbor and receives $\mat{O_{1,1}}$ from the right neighbor. Finally, $\mat{O_{1,1}}$ is
    combined with $\mat{O'_{1,1}}$ (from step (B)) using \attnfix to obtain the final result. The
    output shard $\mat{O_{1,1}}$ is correctly distributed in a 1D cyclic fashion, requiring no
    further redistribution.
    }
\end{figure}

In Subsection~\ref{sec:basic_attn}, we introduced an algorithm to efficiently compute the
self-attention operation using 2D parallelism. In Alg.~\ref{alg:uattn_fwd}, the communication and
computation occurred in separate phases. In this subsection, we enhance this approach by overlapping
computation and communication. This is achieved by performing computation in a block-by-block manner
and using double-buffering to asynchronously send and receive data in the background.
Alg.~\ref{alg:uattn_fwd} consisted of three distinct phases: (1) all-gathering the entire inputs,
(2) performing the flash-attention operation on the gathered inputs, and (3) reduce-scattering the
partial results. By tiling and interleaving these three steps, we can effectively overlap
communication and computation.  This process (from the perspective of a processor $p_{1,1}$) is
illustrated in Figure~\ref{fig:attn2d}, with 16 devices arranged in a $4\times 4$ logical 2D mesh.

\begin{algorithm}
    \begin{algorithmic}[1]
    \Require
    $\mat{Q_p}, \mat{K_p}, \mat{V_p} \in \mathbb{R}^{(N/P)\times H}$: Cyclically distributed Query,
        key, and value matrices;\\
    $p_{r,c}$: Current processor; $P$: Total number of processors

    \State $\mat{K_t}, \mat{V_t} \gets$ \fn{SendRecv}($\set{\mat{K_p}, \mat{V_p}}, p_{c,r}, p_{c,r}$)
    \Comment{Column $\rightarrow$ row-major distribution} \label{alg:attn_fwd:send_kv}

    \State $l \gets (c + \rootp - 1) \% \rootp$,~~~$g \gets (c+1)\%\rootp$
    \Comment{Left and right processor column ids}

    \State $\mat{Q_n}, h_g \gets $ \fn{AsyncSendRecv}($\mat{Q_p}, p_{r,l}, p_{r,g}$)
        \label{alg:attn_fwd:gather_first_block}
        \Comment{Send first block of $\mat{Q}$ in background}

    \State $\mat{K_g}, \mat{V_g}, M_1, \mat{N_1}, D_1 \gets \gc(\mat{Q_p}, \mat{K_t},
        \mat{V_t}$, $p_{r,c}$, $P$)
        \label{alg:attn_fwd:gc}
        \Comment{Gather+compute -- Alg.~\ref{alg:gc}}

    \State $\fn{Wait}(h_g)$ \Comment{Wait for the gather of $\mat{Q_n}$ to finish}
    \State $M_2, \mat{N_2}, D_2 \gets \gcs(\mat{Q_n}, \mat{K_g}, \mat{V_g}$, $p_{r, c}$, $P$)
        \label{alg:attn_fwd:gcs}
        \Comment{Gather+compute+scatter -- Alg.~\ref{alg:gcs}}

    \State $M_p, \mat{N_p}, D_p \gets \attnfix(M_1, \mat{N_1}, D_1, M_2,
        \mat{N_2}, D_2)$ \label{alg:attn_fwd:reduce} \Comment{Perform final reduction}
    \State $\mat{O_p} \gets \fn{Diag}(D_p^{-1})\mat{N_p}$
    \label{alg:attn_fwd:O_p}
    \State \fn{SaveForBackprop}($\mat{Q_p}, \mat{K_t}, \mat{V_t}, M_p, D_p, \mat{O_p}$)
    \State \Return $\mat{O_p}$
\end{algorithmic}
    \caption{\label{alg:attn_fwd}\fn{\attn}: Forward pass of 2D parallel self-attention with overlapping
    communication.}
\end{algorithm}

Alg.~\ref{alg:attn_fwd} outlines the pseudocode for performing the forward pass of a 2D parallel
self-attention operation, incorporating overlapping computation and communication. Initially, the
$\mat{K}$ and $\mat{V}$ matrices are transformed from column-major to row-major distribution format
in line~\ref{alg:attn_fwd:send_kv}.
In line~\ref{alg:attn_fwd:gather_first_block}, the function $\fn{AsyncSendRecv}$ is used to
asynchronously send each processor's block of query matrix, $\mat{Q_p}$, to the processor on its
left ($p_{r,l}$) and receive the next block from the processor on its right ($p_{r,g}$). The
function $\fn{AsyncSendRecv}$ immediately returns a handle $h_g$, which the processor can later use
to wait for the communication to complete.
Concurrently, each processor starts a overlapping gather-compute operation in
line~\ref{alg:attn_fwd:gc} using the \gc function.
Alg.~\ref{alg:gc} provides the pseudocode for \gc, detailing the following steps (refer to
Figure~\ref{fig:attn2d}(B)):
(1) Each processor asynchronously sends its block of key and value matrices, $\mat{K_i}, \mat{V_i}
\in \rspace{N/P \times H}$, to its top neighbor (line~\ref{alg:gc:gather} of Alg.~\ref{alg:gc});
(2) Concurrently, it performs self-attention computation on $\mat{Q_p}$, $\mat{K_i}$, and
$\mat{V_i}$ to generate partial results (line~\ref{alg:gc:compute} of Alg.~\ref{alg:gc});
(3) These partial results are locally reduced with those from previous iteration using \attnfix
(line~\ref{alg:gc:reduce} of Alg.~\ref{alg:gc}).

\begin{algorithm}
\begin{algorithmic}[1]
    \Require
    $\mat{Q_p}, \mat{K_1}, \mat{V_1} \in \mathbb{R}^{(N/P)\times H}$: Query, key, and value
    matrices;\\
    $p_{r,c}$: Current processor; $P$: Total number of processors

    \State $u \gets (r + \rootp - 1) \% \rootp$, $d \gets (r+1)\%\rootp$
    \Comment{Up and down processor row ids}
    \For{$i \gets 1$ to $\rootp$}
    \IfThen{$i > 1$}{$\fn{Wait}(h_g)$} \Comment{Wait for gather data to arrive}
    \IfThen{$i < \rootp$}{$\mat{K_{i+1}}, \mat{V_{i+1}}, h_g \gets
    \fn{AsyncSendRecv}(\set{\mat{K_i}, \mat{V_i}}, p_{u,c}, p_{d,c})$}
    \label{alg:gc:gather} \Comment{Gather}

    \State $M_i, \mat{N_i}, D_i \gets \flashattnfwd(\mat{Q_p}, \mat{K_i}, \mat{V_i})$ 
    \label{alg:gc:compute}\Comment{Compute}
    \If{$i > 1$}
    \State $M, \mat{N}, D \gets \attnfix(M_i, \mat{N_i}, D_i, M, \mat{N}, D)$
    \label{alg:gc:reduce}\Comment{Reduce}
    \Else
    \State $M, \mat{N}, D \gets M_1, \mat{N_1}, D_1$
    \EndIf
    \EndFor
    \State \Return $\fn{Concat}(\mat{K_1},\dots,\mat{K_{\rootp}}),
    \fn{Concat}(\mat{V_1},\dots,\mat{V_{\rootp}}), M, \mat{N}, D$
\end{algorithmic}
    \caption{\label{alg:gc}\gc: Overlapping all-gather and compute operation
    (with local reduction).}
\end{algorithm}

The output of the \gc function in line~\ref{alg:attn_fwd:gc} includes the key and value matrices
$\mat{K_g}, \mat{V_g} \in \rspace{N/\rootp \times H}$, which contain the tokens corresponding to
indices $\set{c + i\rootp~|~i \in [0,N/\rootp)}$. Additionally, it returns the self-attention
outputs $M_1, D_1 \in \rspace{N/P}$ and $\mat{N_1} \in \rspace{N/P \times H}$, resulting from the
self-attention computation on a single query block $\mat{Q_p}$ and the gathered key and value blocks
$\mat{K_g}$ and $\mat{V_g}$.
At this stage, all processors possess the necessary key and value tokens ($\mat{K_g}$ and
$\mat{V_g}$) to carry out their local computations, but they do not have the complete set of query
tokens yet.
The function \gcs is invoked in line~\ref{alg:attn_fwd:gcs} of Alg.~\ref{alg:attn_fwd} to continue
the computation, overlapping the computation with the gathering of the remaining query blocks and
the reduce-scattering of the computed partial results.
Alg.~\ref{alg:gcs} provides the pseudocode for \gcs, which performs the
following operations (refer to Figure~\ref{fig:attn2d}(D)):
(1) Each processor asynchronously sends its block of the query matrix $\mat{Q_n} \in \rspace{N/P \times
H}$ to its left neighbor (line~\ref{alg:gcs:gather} of Alg.~\ref{alg:gcs});
(2) Concurrently, it computes self-attention on $\mat{Q_n}$ and the previously gathered $\mat{K_g},
\mat{V_g} \in \rspace{N/\rootp \times H}$ (line~\ref{alg:gcs:compute} of Alg.~\ref{alg:gcs});
(3) Upon completing the computation, the processor waits for the arrival of partial
results from the right neighbor, then reduces the received results with its local results
(line~\ref{alg:gcs:reduce} of Alg.~\ref{alg:gcs});
(4) This reduced result is then asynchronously passed to the left neighbor
(line~\ref{alg:gcs:scatter} of Alg.~\ref{alg:gcs})). \gcs employs a double-buffer technique to keep
only two query blocks active at any iteration, and discarding older blocks that have already been
used for computation.

\begin{algorithm}
\begin{algorithmic}[1]
    \Require
    $\mat{Q_n} \in \mathbb{R}^{(N/P)\times H}, \mat{K_g}, \mat{V_g} \in \mathbb{R}^{(N/\rootp)\times H}$:
    Query, key, and value matrices;\\
    $p_{r,c}$: Current processor; $P$: Total number of processors

    \State $l \gets (c + \rootp - 1) \% \rootp$, $g \gets (c+1)\%\rootp$
    \Comment{Left and right processor column ids}

    \For{$i \gets 1$ to $\rootp-1$}
    \IfThen{$i > 1$}{$\fn{Wait}(h_g)$} \Comment{Wait for gather data to arrive}
    \IfThen{$i < \rootp-1$}{$\mat{Q_b}, h_g \gets \fn{AsyncSendRecv}(\mat{Q_n}, p_{r,l},
    p_{r,g})$} \label{alg:gcs:gather} \Comment{Gather}

    \State $M_p, \mat{N_p}, D_p \gets \flashattnfwd(\mat{Q_n}, \mat{K_g}, \mat{V_g})$
    \label{alg:gcs:compute} \Comment{Compute}
    \If{$i > 1$}
    \State $\fn{Wait}(h_s)$ \label{alg:gcs:wait_hs} \Comment{Wait for previous scatter data to arrive}
    \State $M_p, \mat{N_p}, D_p \gets \attnfix(M_p, \mat{N_p}, D_p, M, \mat{N}, D)$
    \label{alg:gcs:reduce} \Comment{Reduce}
    \EndIf

    \State $M, \mat{N}, D, h_s \gets \fn{AsyncSendRecv}(\set{M_p, \mat{N_p}, D_p}, p_{r,l}, p_{r,g})$
    \label{alg:gcs:scatter} \Comment{Scatter}

    \State $\mat{Q_n}, \mat{Q_b} \gets \mat{Q_b}, \mat{Q_n}$ \Comment{Swap gather double buffer pointers}

    \EndFor
    \State $\fn{Wait}(h_s)$ \Comment{Wait for final scatter data to arrive}
    \State \Return $M, \mat{N}, D$
\end{algorithmic}
    \caption{\label{alg:gcs}\gcs: Overlapping all-gather, compute and reduce-scatter operation.
    Double buffering is used to eliminate the need to store the entire gathered $\mat{Q_g} \in
    \rspace{N/\rootp \times H}$ matrix.}
\end{algorithm}

The outputs $M_2$, $\mat{N_2}$, and $D_2$ returned by \gcs in line~\ref{alg:attn_fwd:gcs} of
Alg.~\ref{alg:attn_fwd} contain
partial results that are correctly distributed, \ie, each processor $p_{r,c}$ holds the outputs
corresponding to the correct set of token indices $\set{r + \rootp c + iP~|~i \in [0,N/P)}$
for a 1D cyclic distribution.
Additionally, these $M_2$, $\mat{N_2}$, and $D_2$ already include the partial
results from processors $p_{i,c}$ for $i \in [0,\rootp) \setminus r$, which have been reduced
together. Therefore, the local partial results $M_1$, $\mat{N_1}$, and $D_1$ of the current processor
from line~\ref{alg:attn_fwd:gc} are finally reduced with $M_2$, $\mat{N_2}$, and $D_2$ in
line~\ref{alg:attn_fwd:reduce} to obtain the complete results $M_p$, $\mat{N_p}$, and $D_p$. The
gathered key and value matrices $\mat{K_g}$ and $\mat{V_g}$ are discarded, while the query, key and
value blocks $\mat{Q_p}, \mat{K_t}, \mat{V_t} \in \rspace{N/P \times H}$ are saved for the backward
pass. The backward pass is computed in a similar fashion, with overlapping computation and
communication. Details of the backward pass can be found in Appendix~\ref{sec:tiled_attn_bwd}.

\section{\label{sec:results}Results}
We integrated both non-overlapping \basicattn and overlapping \attn parallelization methods into the
Megatron-LM framework~\citep{megatronlm_github} and evaluated them against Nvidia Transformer
Engine's~\citep{transformerengine} efficient implementation of Ring
Attention~\citep{liu2023ringattentionblockwisetransformers}, also known as Context Parallelism.
The evaluation was performed with the GPT-3-like model architecture~\citep{gpt3, opt}, featuring parameters
of 760M, 2.7B, 13B, 66B and 175B. Details of the model architecture are provided in Table~\ref{tbl:gpt3arch}.
GPT-3 is a causal language model. The original Ring
Attention's~\citep{liu2023ringattentionblockwisetransformers} block distribution caused load imbalances when applying
causal attention. However, Transformer Engine's~\citep{transformerengine} implementation of Ring
Attention efficiently addresses this issue by reordering the input sequence tokens as follows: The tokens are
divided into two equal halves along the sequence length dimension. The first half is block
distributed equally among all processors from $p_0$ to $p_{P-1}$, and the second half is block
distributed equally among all processors in reverse order, from $p_{P-1}$ to $p_0$.  This results in
a fully load-balanced computation. In our \basicattn and \attn methods, we use a cyclic distribution
instead of block distribution, which inherently balances the computation load.

\begin{table}
    \centering
    \caption{\label{tbl:gpt3arch}Specifications of model architectures employed in our experiments.}
    \begin{tabular}{lcccc}
        \toprule
        \textbf{Parameters} & \textbf{Layers ($L$)} & \textbf{Heads ($M$)} & \textbf{Head size ($H$)}
        & \textbf{Hidden size ($M \times H$)} \\
        \midrule
        760M & 24 & 16 & 96 & 1536 \\
        2.7B & 32 & 32 & 80 & 2560 \\
        13B & 40 & 40 & 128 & 5120 \\
        66B & 64 & 72 & 128 & 9216 \\
        175B & 96 & 96 & 128 & 12288 \\
        \bottomrule
    \end{tabular}
\end{table}

\subsection{\label{sec:results:a100}Multi-node Speedup with A100 GPUs}
The experiments were carried out on a multi-node/multi-GPU configuration, with each node equipped
with 4 NVIDIA A100 GPUs connected via PCIe, and inter-node communication facilitated through an
Ethernet network.
We used global batch sizes of $64$, $32$, and $16$ for model sizes of 760M, 2.7B, and 13B, respectively.
Sequence lengths were varied from $2^{13}=8192$ to $2^{17}=131072$, increasing in powers of 2.
We configured the degrees of pipeline, data and tensor parallelism to 1, and allocated all available GPUs to
context parallelism for both Ring Attention and our 2D parallel methods during our experiments.
This configuration allows us to compare our method directly with Ring Attention, without the
influence of pipeline and data parallelism on the comparison.
In practice, depending on the heterogeneity of the distributed setup and the available memory on
each device, context parallelism can be combined with pipeline parallelism to achieve optimal
performance.
We enabled ZeRO~\citep{zero} parallelism for non-self-attention layers such as FFN and
QKV-projections in our experiments.
We did not use CPU offloading.
Additionally, we disabled layerwise activation checkpointing\footnote{Disabling layerwise activation
checkpointing turns off the activation checkpointing feature of
Megatron-LM~\citep{megatronlm_github} code. However, the activation checkpointing for \flashattnfwd,
which is essential to avoid storing the $\bigo{N^2}$ attention score matrix for backpropagation,
remains enabled.}
for all our runs, except for the 13B model with sequence lengths of $2^{16}$ and $2^{17}$, where
activation checkpointing was necessary to prevent the model from exhausting GPU memory.

\begin{figure}
    \centering
    \includegraphics[width=\textwidth]{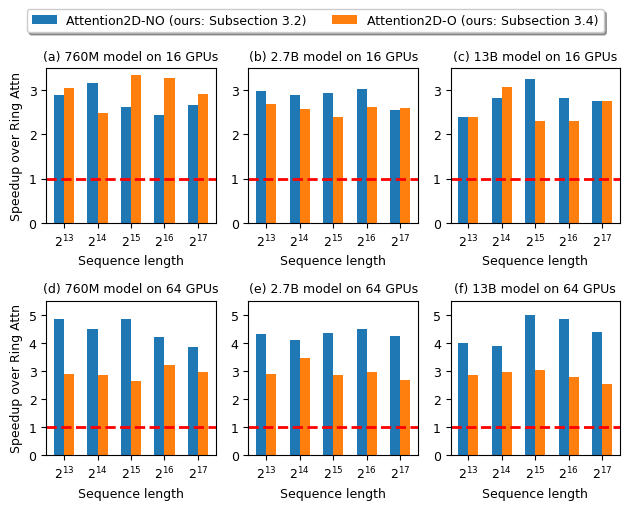}
    \caption{\label{fig:speedup}Training performance improvement of \basicattn
    (Subsection~\ref{sec:basic_attn}) and \attn (Subsection~\ref{sec:tiled_attn}) relative to Ring
    Attention across various model sizes on 16 GPUs (distributed over 4 nodes) and 64 GPUs
    (distributed over 16 nodes).}
\end{figure}

Figure~\ref{fig:speedup} presents the speedup of our technique in comparison to Ring
Attention across different model sizes, with $16$ GPUs distributed over $4$ nodes
(Figure~\ref{fig:speedup}(a) -- (c)) and $64$ GPUs distributed over $16$ nodes
(Figure~\ref{fig:speedup}(d) -- (f)). Both our
\basicattn and \attn consistently deliver a speedup of more than twice that of Ring Attention for
all three model sizes and across all sequence lengths.
Specifically, \basicattn delivers a speedup of $\gex{2.4}$ on $16$ GPUs and $\gex{4}$ on $64$ GPUs, while \attn
achieves a speedup of $\gex{2.3}$ on $16$ GPUs and $\gex{2.5}$ on $64$ GPUs.
Interestingly, \attn, which overlaps communication with computation, does not always outperform its
non-overlapping counterpart, \basicattn.
We believe this could be due to two main reasons:
(1) Inefficiencies in our implementation of \attn algorithm. Specifically, instead of performing the
\flashattnfwd operation of a single query block against gathered keys and values in one go as
outlined in line~\ref{alg:gcs:compute} of Alg.~\ref{alg:gcs}, we had to execute this in multiple
iterations, with each query block computed against a single key-value block per iteration. This is
because the current
implementation\footnote{https://github.com/Dao-AILab/flash-attention/tree/v2.4.2} of the
\flashattnfwd kernel~\citep{dao2023flashattention2} only
supports a lower triangular matrix pattern for causal masks. However, the mask required for
self-attention computation of a single query block against gathered keys and values has a step-like
pattern, unlike the lower triangular matrix pattern involved in \basicattn. We plan to address
these kernel-level optimizations in future work.
(2) In \attn, we use a ring-based communication pattern for collective communication operations.
While this pattern is independent of network topology and adapts well to various network
configurations, it doesn't leverage potential optimizations available for the underlying topology.
On the other hand, \basicattn takes advantage of the efficient, topology-aware implementation of
all-gather collective provided by standard libraries like NCCL~\citep{nvidiadoubletree}. We discuss
this topic more in Section~\ref{sec:discussion} and intend to investigate network-aware
communication patterns in future research.

\begin{figure}
    \centering
    \includegraphics[width=0.7\textwidth]{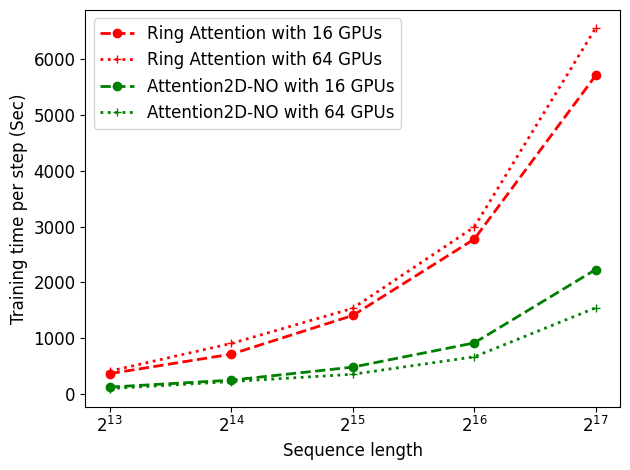}
    \caption{\label{fig:scaling}Training time comparison of Ring Attention and
    \basicattn(Subsection~\ref{sec:basic_attn}) for the 2.7B model, measured with 16 and 64 GPUs
    across various sequence lengths. Training time with our \basicattn decreases effectively as the
    number of processors increases from $16$ to $64$, while training time with Ring Attention not
    only shows no improvement but actually worsens due to additional overheads.}
\end{figure}

\subsection{\label{subsec:results:scaling}Scaling Study}
Figure~\ref{fig:scaling} shows the training time per step for both Ring Attention and our \basicattn
methods for 2.7B model across various sequence lengths.
With Ring Attention, increasing the number of GPUs from $16$ to $64$ does not improve training time; in
fact, the training time increases due to additional overheads, and the slowdown worsens as the
sequence length grows.
Conversely, with our \basicattn, the training time decreases as the number of GPUs increases
from $16$ to $64$, and this improvement becomes more pronounced with longer sequence lengths. This
confirms our assertion that our approach scales effectively with the addition of more processors,
unlike Ring Attention.
As shown in Subsection~\ref{sec:cost_analysis}, \attntext is theoretically expected to scale by
a factor of $\rootp$. Therefore, increasing from $16$ to $64$ GPUs should result in a reduction in
training time of approximately $2\times$. As seen in Figure~\ref{fig:scaling}, for a sequence length
of $131072$, we observe an actual improvement of $1.4\times$.

\subsection{\label{subsec:results:cp_pp}Evaluating Context and Pipeline Parallelism Together}
\begin{figure}
    \centering
    \includegraphics[width=\textwidth]{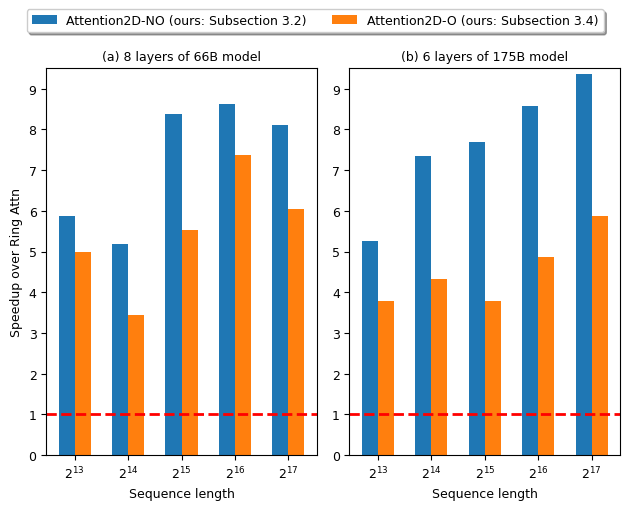}
    \caption{\label{fig:speedup_h100}Training performance improvement of \basicattn
    (Subsection~\ref{sec:basic_attn}) and \attn (Subsection~\ref{sec:tiled_attn}) relative to Ring
    Attention, for a single pipeline stage executed on 64 Ethernet-connected nodes, each equipped
    with a single NVIDIA H100 GPU.}
\end{figure}

It is common practice to use thousands of GPUs for training large language models (LLMs) and to
integrate various parallelism strategies to fit the model into the available memory and optimize
efficiency. For instance, the OPT-175B model~\citep{opt} was trained using 992 80GB A100 GPUs,
employing a combination of Fully Sharded Data Parallel~\citep{fsdp} and Megatron-LM Tensor
Parallelism~\citep{MegatronLM}. In this section, we explore the impact of combining context
parallelism with pipeline parallelism on large-scale billion-parameter LLMs, specifically those with
66B and 175B parameters. Due to the lack of access to thousands of GPUs for a full-scale evaluation,
we simulate only a single stage of pipeline parallelism in our experiments to assess the speedup of
our \attntext method compared to Ring Attention in a single pipeline stage.
For our experiments, we utilize $64$ nodes, each equipped with a single NVIDIA H100 GPU and connected
via an Ethernet network. We fix the number of GPUs assigned for context parallelism to $64$ and
allocate as many model layers as the GPU memory allows for a single pipeline stage. This
configuration enables us to fit $8$ layers of the 66B model within 64 GPUs, resulting in 8 pipeline
stages when the model is trained on $512$ GPUs. Similarly, we can fit $6$ layers of the 175B model
within $64$ GPUs, leading to $16$ pipeline stages when the model is trained on $1024$ GPUs. We used
a global batch size of $16$ for both the models and at all sequence lengths.
Figure~\ref{fig:speedup_h100} shows the speedup of our technique compared to the Ring Attention
technique for a single pipeline stage.
Both our \basicattn and \attn consistently outperform Ring Attention across all sequence
lengths, delivering speedups of up to $8.6\times$ on the 66B model and $9.4\times$ on the 175B
model.

\section{\label{sec:discussion}Discussion and Related Work}
\paragraph{Memory-efficient attention.}
Self-attention is a fundamental component of Transformer models, but it tends to be slow and
memory-intensive, particularly for long sequences, due to its quadratic memory cost relative to
sequence length. This challenge has spurred the creation of memory-efficient attention
mechanisms~\citep{online_softmax, memory_efficient_self_attn, dao2022flashattention,
dao2023flashattention2} that achieve linear memory costs. These methods primarily address memory
cost improvements on a single GPU. Our proposed solution leverages these techniques to reduce memory
costs on a single GPU while introducing an efficient parallelization strategy to scale
self-attention computations across multiple GPUs.

\paragraph{Parallelism methods.}
Various self-attention parallelism methods~\citep{MegatronLM, megatron_seqence_parallelism,
li-etal-2023-sequence, liu2023ringattentionblockwisetransformers, li2023lightseq,
jacobs2023deepspeed, fang2024unified} have been proposed in the past. However, these methods mainly
focus on reducing memory costs, often neglecting communication costs.  As shown in
Table~\ref{tbl:costs}, all these techniques incur a communication cost that increases
linearly with the sequence length, regardless of the number of devices used. In contrast, our
proposed parallelism method achieves a sub-linear communication cost that decreases as the number of
devices increases.
Parallelization strategies like data parallelism~\citep{data_parallelism}, pipeline
parallelism~\citep{narayanan2021memoryefficientpipelineparalleldnntraining}, and
ZeRO~\citep{zero,fsdp} focus on parallelizing the entire LLM.
Many of these methods complement our approach. Therefore, similar to the
strategy employed by USP~\citep{fang2024unified}, a hybrid approach that combines these techniques
with ours can be utilized to achieve further improvements.

\paragraph{Approximation of attention.}
To reduce the computational cost of attention on long contexts, numerous methods have explored
approximations of attention mechanisms~\citep{longformer, performer, reformer,
linformer, transformersrnnsfastautoregressive, bigbird}.
Although these methods have been used in some cases, standard attention is still often preferred
because these approximations can yield sub-optimal results. In contrast, our method does not rely on
any approximations. Instead, we propose improvements to the standard self-attention operation for
parallel settings.
Grouped Query Attention
(GQA)~\citep{gqa} and Multi-Query Attention (MQA)~\cite{mqa} are variants of standard self-attention
where multiple query heads attend to the same key and value head, reducing the size of the KV cache
during inference. Since our parallelism method operates at the level of single attention heads, it
can be directly applied to these variants.

\paragraph{Long context inference.}
In Subsection~\ref{sec:parallel_kv}, we applied the associative property within the Softmax operation
to reveal parallelism along the \kvdim. This concept of leveraging the associative property to
break the dependence along the \kvdim has also been recognized by others in the context of long
context length inference~\citep{flashdecoding, leanattn, treeattn}.
These methods primarily aim to enhance inference latency during the token-generation phase of
transformer models, where the query length is typically $1$. Conversely, our focus is on improving
training throughput and inference latency during the prompt phase, where the query length equals the
key/value length, $N$.

\paragraph{Network topology-aware collectives.}
In Subsection~\ref{sec:tiled_attn}, we developed an efficient self-attention mechanism that overlaps
computation and communication. We employed a ring-based communication pattern to gather inputs and
reduce-scatter outputs. Although ring-based collective communication is independent of topology and
generalizes well across various networks, specialized topology-aware collective algorithms can offer
additional benefits. Several such collective communication algorithms have been developed to reduce
communication costs for specific network
topologies~\citep{nvidiadoubletree, jia2018highlyscalabledeeplearning,
mikami2019massivelydistributedsgdimagenetresnet50, ying2018imageclassificationsupercomputerscale}.
We plan to explore these possibilities in our future work.

\section{\label{sec:conclusion}Conclusion}
The self-attention operation, a fundamental component of Transformer models, is both memory and
communication intensive. While advancements have reduced the memory cost from quadratic to linear
with respect to context length, progress in reducing communication cost has been minimal. The
communication cost of the current state-of-the-art technique remains linear in context length,
regardless of the number of processors used. In this paper, we introduced a novel parallelism
scheme, \attntext, that is asymptotically faster than previous methods. It achieves a sub-linear
communication cost that decreases with an increasing number of processors, without requiring
approximations or incurring additional computational or memory overheads. Our experiments
demonstrated that our approach achieved up to a $5\times$ performance
improvement on a multi-node/multi-GPU setup with NVIDIA A100 GPUs, and up to a $9.4\times$
performance improvement on a multi-node/single-GPU setup with NVIDIA H100 GPUs, outperforming the
current state-of-the-art across different model sizes and sequence lengths. Additionally, our
experiments confirmed that, unlike previous methods, our proposed method scaled effectively with an
increasing number of processors.

\section*{Acknowledgment}
We thank Partha Maji for their valuable feedback and suggestions, which helped improve the paper.

{\small
\bibliography{attn2d}

\begin{thebibliography}{38}
\providecommand{\natexlab}[1]{#1}
\providecommand{\url}[1]{\texttt{#1}}
\expandafter\ifx\csname urlstyle\endcsname\relax
  \providecommand{\doi}[1]{doi: #1}\else
  \providecommand{\doi}{doi: \begingroup \urlstyle{rm}\Url}\fi

\bibitem[Ainslie et~al.(2023)Ainslie, Lee-Thorp, de~Jong, Zemlyanskiy, Lebrón,
  and Sanghai]{gqa}
Joshua Ainslie, James Lee-Thorp, Michiel de~Jong, Yury Zemlyanskiy, Federico
  Lebrón, and Sumit Sanghai.
\newblock Gqa: Training generalized multi-query transformer models from
  multi-head checkpoints, 2023.
\newblock URL \url{https://arxiv.org/abs/2305.13245}.

\bibitem[Anthropic(2024)]{claude3modelcard2024}
Anthropic.
\newblock Claude 3 model card, 2024.
\newblock URL
  \url{https://www-cdn.anthropic.com/f2986af8d052f26236f6251da62d16172cfabd6e/claude-3-model-card.pdf}.

\bibitem[Artetxe et~al.(2022)Artetxe, Bhosale, Goyal, Mihaylov, Ott, Shleifer,
  Lin, Du, Iyer, Pasunuru, Anantharaman, Li, Chen, Akin, Baines, Martin, Zhou,
  Koura, O'Horo, Wang, Zettlemoyer, Diab, Kozareva, and Stoyanov]{fsdp}
Mikel Artetxe, Shruti Bhosale, Naman Goyal, Todor Mihaylov, Myle Ott, Sam
  Shleifer, Xi~Victoria Lin, Jingfei Du, Srinivasan Iyer, Ramakanth Pasunuru,
  Giri Anantharaman, Xian Li, Shuohui Chen, Halil Akin, Mandeep Baines, Louis
  Martin, Xing Zhou, Punit~Singh Koura, Brian O'Horo, Jeff Wang, Luke
  Zettlemoyer, Mona Diab, Zornitsa Kozareva, and Ves Stoyanov.
\newblock Efficient large scale language modeling with mixtures of experts,
  2022.
\newblock URL \url{https://arxiv.org/abs/2112.10684}.

\bibitem[Beltagy et~al.(2020)Beltagy, Peters, and Cohan]{longformer}
Iz~Beltagy, Matthew~E. Peters, and Arman Cohan.
\newblock Longformer: The long-document transformer, 2020.
\newblock URL \url{https://arxiv.org/abs/2004.05150}.

\bibitem[Brown et~al.(2020)Brown, Mann, Ryder, Subbiah, Kaplan, Dhariwal,
  Neelakantan, Shyam, Sastry, Askell, Agarwal, Herbert-Voss, Krueger, Henighan,
  Child, Ramesh, Ziegler, Wu, Winter, Hesse, Chen, Sigler, Litwin, Gray, Chess,
  Clark, Berner, McCandlish, Radford, Sutskever, and Amodei]{gpt3}
Tom Brown, Benjamin Mann, Nick Ryder, Melanie Subbiah, Jared~D Kaplan, Prafulla
  Dhariwal, Arvind Neelakantan, Pranav Shyam, Girish Sastry, Amanda Askell,
  Sandhini Agarwal, Ariel Herbert-Voss, Gretchen Krueger, Tom Henighan, Rewon
  Child, Aditya Ramesh, Daniel Ziegler, Jeffrey Wu, Clemens Winter, Chris
  Hesse, Mark Chen, Eric Sigler, Mateusz Litwin, Scott Gray, Benjamin Chess,
  Jack Clark, Christopher Berner, Sam McCandlish, Alec Radford, Ilya Sutskever,
  and Dario Amodei.
\newblock Language models are few-shot learners.
\newblock In H.~Larochelle, M.~Ranzato, R.~Hadsell, M.F. Balcan, and H.~Lin,
  editors, \emph{Advances in Neural Information Processing Systems}, volume~33,
  pages 1877--1901. Curran Associates, Inc., 2020.
\newblock URL
  \url{https://proceedings.neurips.cc/paper_files/paper/2020/file/1457c0d6bfcb4967418bfb8ac142f64a-Paper.pdf}.

\bibitem[Chan et~al.(2007)Chan, Heimlich, Purkayastha, and van~de
  Geijn]{collective_comm}
Ernie Chan, Marcel Heimlich, Avi Purkayastha, and Robert van~de Geijn.
\newblock Collective communication: theory, practice, and experience: Research
  articles.
\newblock \emph{Concurrency and Computation: Practice and Experience},
  19\penalty0 (13):\penalty0 1749–1783, sep 2007.
\newblock ISSN 1532-0626.

\bibitem[Choromanski et~al.(2022)Choromanski, Likhosherstov, Dohan, Song, Gane,
  Sarlos, Hawkins, Davis, Mohiuddin, Kaiser, Belanger, Colwell, and
  Weller]{performer}
Krzysztof Choromanski, Valerii Likhosherstov, David Dohan, Xingyou Song,
  Andreea Gane, Tamas Sarlos, Peter Hawkins, Jared Davis, Afroz Mohiuddin,
  Lukasz Kaiser, David Belanger, Lucy Colwell, and Adrian Weller.
\newblock Rethinking attention with performers, 2022.
\newblock URL \url{https://arxiv.org/abs/2009.14794}.

\bibitem[Dao(2024)]{dao2023flashattention2}
Tri Dao.
\newblock Flash{A}ttention-2: Faster attention with better parallelism and work
  partitioning.
\newblock In \emph{International Conference on Learning Representations
  (ICLR)}, 2024.

\bibitem[Dao et~al.(2022)Dao, Fu, Ermon, Rudra, and
  R{\'e}]{dao2022flashattention}
Tri Dao, Daniel~Y. Fu, Stefano Ermon, Atri Rudra, and Christopher R{\'e}.
\newblock Flash{A}ttention: Fast and memory-efficient exact attention with
  {IO}-awareness.
\newblock In \emph{Advances in Neural Information Processing Systems
  (NeurIPS)}, 2022.

\bibitem[Dao et~al.(2023)Dao, Haziza, Massa, and Sizov]{flashdecoding}
Tri Dao, Daniel Haziza, Francisco Massa, and Grigory Sizov.
\newblock {F}lash-{D}ecoding, 2023.
\newblock URL \url{https://crfm.stanford.edu/2023/10/12/flashdecoding.html}.

\bibitem[Dean et~al.(2012)Dean, Corrado, Monga, Chen, Devin, Mao, Ranzato,
  Senior, Tucker, Yang, Le, and Ng]{data_parallelism}
Jeffrey Dean, Greg Corrado, Rajat Monga, Kai Chen, Matthieu Devin, Mark Mao,
  Marc\textquotesingle~aurelio Ranzato, Andrew Senior, Paul Tucker, Ke~Yang,
  Quoc Le, and Andrew Ng.
\newblock Large scale distributed deep networks.
\newblock In F.~Pereira, C.J. Burges, L.~Bottou, and K.Q. Weinberger, editors,
  \emph{Advances in Neural Information Processing Systems}, volume~25. Curran
  Associates, Inc., 2012.
\newblock URL
  \url{https://proceedings.neurips.cc/paper_files/paper/2012/file/6aca97005c68f1206823815f66102863-Paper.pdf}.

\bibitem[Fang and Zhao(2024)]{fang2024unified}
Jiarui Fang and Shangchun Zhao.
\newblock Usp: A unified sequence parallelism approach for long context
  generative ai.
\newblock \emph{arXiv preprint arXiv:2405.07719}, 2024.

\bibitem[Jacobs et~al.(2023)Jacobs, Tanaka, Zhang, Zhang, Song, Rajbhandari,
  and He]{jacobs2023deepspeed}
Sam~Ade Jacobs, Masahiro Tanaka, Chengming Zhang, Minjia Zhang, Shuaiwen~Leon
  Song, Samyam Rajbhandari, and Yuxiong He.
\newblock Deepspeed ulysses: System optimizations for enabling training of
  extreme long sequence transformer models.
\newblock \emph{arXiv preprint arXiv:2309.14509}, 2023.

\bibitem[Jia et~al.(2018)Jia, Song, He, Wang, Rong, Zhou, Xie, Guo, Yang, Yu,
  Chen, Hu, Shi, and Chu]{jia2018highlyscalabledeeplearning}
Xianyan Jia, Shutao Song, Wei He, Yangzihao Wang, Haidong Rong, Feihu Zhou,
  Liqiang Xie, Zhenyu Guo, Yuanzhou Yang, Liwei Yu, Tiegang Chen, Guangxiao Hu,
  Shaohuai Shi, and Xiaowen Chu.
\newblock Highly scalable deep learning training system with mixed-precision:
  Training imagenet in four minutes, 2018.
\newblock URL \url{https://arxiv.org/abs/1807.11205}.

\bibitem[Katharopoulos et~al.(2020)Katharopoulos, Vyas, Pappas, and
  Fleuret]{transformersrnnsfastautoregressive}
Angelos Katharopoulos, Apoorv Vyas, Nikolaos Pappas, and François Fleuret.
\newblock Transformers are rnns: Fast autoregressive transformers with linear
  attention, 2020.
\newblock URL \url{https://arxiv.org/abs/2006.16236}.

\bibitem[Kitaev et~al.(2020)Kitaev, Łukasz Kaiser, and Levskaya]{reformer}
Nikita Kitaev, Łukasz Kaiser, and Anselm Levskaya.
\newblock Reformer: The efficient transformer, 2020.
\newblock URL \url{https://arxiv.org/abs/2001.04451}.

\bibitem[Korthikanti et~al.(2022)Korthikanti, Casper, Lym, McAfee, Andersch,
  Shoeybi, and Catanzaro]{megatron_seqence_parallelism}
Vijay Korthikanti, Jared Casper, Sangkug Lym, Lawrence McAfee, Michael
  Andersch, Mohammad Shoeybi, and Bryan Catanzaro.
\newblock Reducing activation recomputation in large transformer models, 2022.
\newblock URL \url{https://arxiv.org/abs/2205.05198}.

\bibitem[Li et~al.(2023{\natexlab{a}})Li, Shao, Xie, Xing, Gonzalez, Stoica,
  Ma, and Zhang]{li2023lightseq}
Dacheng Li, Rulin Shao, Anze Xie, Eric Xing, Joseph Gonzalez, Ion Stoica,
  Xuezhe Ma, and Hao Zhang.
\newblock Lightseq: : Sequence level parallelism for distributed training of
  long context transformers.
\newblock In \emph{Workshop on Advancing Neural Network Training: Computational
  Efficiency, Scalability, and Resource Optimization (WANT@NeurIPS 2023)},
  2023{\natexlab{a}}.
\newblock URL \url{https://openreview.net/forum?id=qScA3fL49l}.

\bibitem[Li et~al.(2023{\natexlab{b}})Li, Xue, Baranwal, Li, and
  You]{li-etal-2023-sequence}
Shenggui Li, Fuzhao Xue, Chaitanya Baranwal, Yongbin Li, and Yang You.
\newblock Sequence parallelism: Long sequence training from system perspective.
\newblock In Anna Rogers, Jordan Boyd-Graber, and Naoaki Okazaki, editors,
  \emph{Proceedings of the 61st Annual Meeting of the Association for
  Computational Linguistics (Volume 1: Long Papers)}, pages 2391--2404,
  Toronto, Canada, July 2023{\natexlab{b}}. Association for Computational
  Linguistics.
\newblock \doi{10.18653/v1/2023.acl-long.134}.
\newblock URL \url{https://aclanthology.org/2023.acl-long.134}.

\bibitem[Liu et~al.(2023)Liu, Zaharia, and
  Abbeel]{liu2023ringattentionblockwisetransformers}
Hao Liu, Matei Zaharia, and Pieter Abbeel.
\newblock Ring attention with blockwise transformers for near-infinite context,
  2023.
\newblock URL \url{https://arxiv.org/abs/2310.01889}.

\bibitem[Mikami et~al.(2019)Mikami, Suganuma, U-chupala, Tanaka, and
  Kageyama]{mikami2019massivelydistributedsgdimagenetresnet50}
Hiroaki Mikami, Hisahiro Suganuma, Pongsakorn U-chupala, Yoshiki Tanaka, and
  Yuichi Kageyama.
\newblock Massively distributed sgd: Imagenet/resnet-50 training in a flash,
  2019.
\newblock URL \url{https://arxiv.org/abs/1811.05233}.

\bibitem[Milakov and Gimelshein(2018)]{online_softmax}
Maxim Milakov and Natalia Gimelshein.
\newblock Online normalizer calculation for softmax.
\newblock \emph{CoRR}, abs/1805.02867, 2018.
\newblock URL \url{http://arxiv.org/abs/1805.02867}.

\bibitem[Narayanan et~al.(2021)Narayanan, Phanishayee, Shi, Chen, and
  Zaharia]{narayanan2021memoryefficientpipelineparalleldnntraining}
Deepak Narayanan, Amar Phanishayee, Kaiyu Shi, Xie Chen, and Matei Zaharia.
\newblock Memory-efficient pipeline-parallel {DNN} training, 2021.
\newblock URL \url{https://arxiv.org/abs/2006.09503}.

\bibitem[NVIDIA({\natexlab{a}})]{megatronlm_github}
NVIDIA.
\newblock {M}egatron-{LM}, {\natexlab{a}}.
\newblock URL \url{https://github.com/NVIDIA/Megatron-LM/tree/core_r0.8.0}.

\bibitem[NVIDIA({\natexlab{b}})]{transformerengine}
NVIDIA.
\newblock {T}ransformer {E}ngine, {\natexlab{b}}.
\newblock URL
  \url{https://github.com/NVIDIA/TransformerEngine/tree/release_v1.9}.

\bibitem[NVIDIA(2019)]{nvidiadoubletree}
NVIDIA.
\newblock Massively scale your deep learning training with {NCCL} 2.4, 2019.
\newblock URL
  \url{https://developer.nvidia.com/blog/massively-scale-deep-learning-training-nccl-2-4/}.

\bibitem[OpenAI(2024)]{gpt4odocs2024}
OpenAI.
\newblock Gpt-4o documentation, 2024.
\newblock URL \url{https://platform.openai.com/docs/models/gpt-4o}.

\bibitem[Rabe and Staats(2021)]{memory_efficient_self_attn}
Markus~N. Rabe and Charles Staats.
\newblock Self-attention does not need {O}(n\({}^{\mbox{2}}\)) memory.
\newblock \emph{CoRR}, abs/2112.05682, 2021.
\newblock URL \url{https://arxiv.org/abs/2112.05682}.

\bibitem[Rajbhandari et~al.(2020)Rajbhandari, Rasley, Ruwase, and He]{zero}
Samyam Rajbhandari, Jeff Rasley, Olatunji Ruwase, and Yuxiong He.
\newblock Zero: memory optimizations toward training trillion parameter models.
\newblock In \emph{Proceedings of the International Conference for High
  Performance Computing, Networking, Storage and Analysis}, SC '20. IEEE Press,
  2020.
\newblock ISBN 9781728199986.

\bibitem[Sanovar et~al.(2024)Sanovar, Bharadwaj, Amant, Rühle, and
  Rajmohan]{leanattn}
Rya Sanovar, Srikant Bharadwaj, Renee~St. Amant, Victor Rühle, and Saravan
  Rajmohan.
\newblock Lean attention: Hardware-aware scalable attention mechanism for the
  decode-phase of transformers, 2024.
\newblock URL \url{https://arxiv.org/abs/2405.10480}.

\bibitem[Shazeer(2019)]{mqa}
Noam Shazeer.
\newblock Fast transformer decoding: One write-head is all you need, 2019.
\newblock URL \url{https://arxiv.org/abs/1911.02150}.

\bibitem[Shoeybi et~al.(2019)Shoeybi, Patwary, Puri, LeGresley, Casper, and
  Catanzaro]{MegatronLM}
Mohammad Shoeybi, Mostofa Patwary, Raul Puri, Patrick LeGresley, Jared Casper,
  and Bryan Catanzaro.
\newblock Megatron-lm: Training multi-billion parameter language models using
  model parallelism.
\newblock \emph{CoRR}, abs/1909.08053, 2019.
\newblock URL \url{http://arxiv.org/abs/1909.08053}.

\bibitem[Shyam et~al.(2024)Shyam, Pilault, Shepperd, Anthony, and
  Millidge]{treeattn}
Vasudev Shyam, Jonathan Pilault, Emily Shepperd, Quentin Anthony, and Beren
  Millidge.
\newblock Tree attention: Topology-aware decoding for long-context attention on
  gpu clusters, 2024.
\newblock URL \url{https://arxiv.org/abs/2408.04093}.

\bibitem[Vaswani et~al.(2017)Vaswani, Shazeer, Parmar, Uszkoreit, Jones, Gomez,
  Kaiser, and Polosukhin]{NIPS2017_3f5ee243}
Ashish Vaswani, Noam Shazeer, Niki Parmar, Jakob Uszkoreit, Llion Jones,
  Aidan~N Gomez, \L~ukasz Kaiser, and Illia Polosukhin.
\newblock Attention is all you need.
\newblock In I.~Guyon, U.~Von Luxburg, S.~Bengio, H.~Wallach, R.~Fergus,
  S.~Vishwanathan, and R.~Garnett, editors, \emph{Advances in Neural
  Information Processing Systems}, volume~30. Curran Associates, Inc., 2017.
\newblock URL
  \url{https://proceedings.neurips.cc/paper_files/paper/2017/file/3f5ee243547dee91fbd053c1c4a845aa-Paper.pdf}.

\bibitem[Wang et~al.(2020)Wang, Li, Khabsa, Fang, and Ma]{linformer}
Sinong Wang, Belinda~Z. Li, Madian Khabsa, Han Fang, and Hao Ma.
\newblock Linformer: Self-attention with linear complexity, 2020.
\newblock URL \url{https://arxiv.org/abs/2006.04768}.

\bibitem[Ying et~al.(2018)Ying, Kumar, Chen, Wang, and
  Cheng]{ying2018imageclassificationsupercomputerscale}
Chris Ying, Sameer Kumar, Dehao Chen, Tao Wang, and Youlong Cheng.
\newblock Image classification at supercomputer scale, 2018.
\newblock URL \url{https://arxiv.org/abs/1811.06992}.

\bibitem[Zaheer et~al.(2021)Zaheer, Guruganesh, Dubey, Ainslie, Alberti,
  Ontanon, Pham, Ravula, Wang, Yang, and Ahmed]{bigbird}
Manzil Zaheer, Guru Guruganesh, Avinava Dubey, Joshua Ainslie, Chris Alberti,
  Santiago Ontanon, Philip Pham, Anirudh Ravula, Qifan Wang, Li~Yang, and Amr
  Ahmed.
\newblock Big bird: Transformers for longer sequences, 2021.
\newblock URL \url{https://arxiv.org/abs/2007.14062}.

\bibitem[Zhang et~al.(2022)Zhang, Roller, Goyal, Artetxe, Chen, Chen, Dewan,
  Diab, Li, Lin, Mihaylov, Ott, Shleifer, Shuster, Simig, Koura, Sridhar, Wang,
  and Zettlemoyer]{opt}
Susan Zhang, Stephen Roller, Naman Goyal, Mikel Artetxe, Moya Chen, Shuohui
  Chen, Christopher Dewan, Mona Diab, Xian Li, Xi~Victoria Lin, Todor Mihaylov,
  Myle Ott, Sam Shleifer, Kurt Shuster, Daniel Simig, Punit~Singh Koura, Anjali
  Sridhar, Tianlu Wang, and Luke Zettlemoyer.
\newblock {OPT}: {O}pen {P}re-trained {T}ransformer {L}anguage {M}odels, 2022.
\newblock URL \url{https://arxiv.org/abs/2205.01068}.

\end{thebibliography}
\bibliographystyle{plainnat}
}

\appendix
\section{\label{sec:row_reduce_scatter}Ring-Based Reduce-Scatter With Cyclic Distribution}
Alg.~\ref{alg:reduce_scatter} outlines a standard ring-based reduce-scatter operation tailored for
cyclic data distribution within the rows of a two-dimensional processor grid. 
In line~\ref{alg:reduce_scatter:neighbors}, the left and right neighbors of the current processor
$p_{r,c}$ are identified. During each step, the current processor sends a chunk of data to its left
neighbor and receives data from its right neighbor.
In line~\ref{alg:reduce_scatter:idx}, an index set $idx$ is generated using a
cyclic distribution with a stride of $\rootp$ and an initial offset of $(c+1)\%\rootp$. This set
corresponds to the indices for the first chunk of data to be sent to the left neighbor.
In line~\ref{alg:reduce_scatter:shard}, shards of $M_g$, $\mat{N_g}$, and $D_g$ are extracted using
the index set $idx$ and sent to the left neighbor in line~\ref{alg:reduce_scatter:sendrecv}.
In line~\ref{alg:reduce_scatter:new_idx}, the next index set is calculated by incrementing the
offset by $1$. 
In line~\ref{alg:reduce_scatter:reduce_op}, the next set of shards is extracted and reduced with the
previously received shards from line~\ref{alg:reduce_scatter:sendrecv} using \attnfix
(Alg.~\ref{alg:reduce_op}) as the reduction operator.
These reduced shards are then sent to the left processor in the next iteration.
After $\rootp-1$ steps, each processor ends up with the correct shard of data, with partial outputs
fully reduced.
For more details on the standard ring-based reduce-scatter operation, refer
to~\citet[Section~6.3.2]{collective_comm}.
\begin{algorithm}
\begin{algorithmic}[1]
    \Require
    $M_g \in \rspace{N/\rootp}, \mat{N_g} \in \rspace{(N/\rootp) \times H}, D_g \in
    \rspace{N/\rootp}$: Partial self-attention output of a processor $p_{r,c}$;\\
    $p_{r,c}$: Current processor; $P$: Total number of processors

    \State $l \gets (c + \rootp - 1) \% \rootp$, $g \gets (c+1)\%\rootp$
    \label{alg:reduce_scatter:neighbors}\Comment{Left and right processor column ids}
    \State $idx \gets \set{(c+1)\%\rootp + j\rootp~|~j\in [0,N/\rootp)}$
    \label{alg:reduce_scatter:idx}\Comment{Cyclic distribution}
    \State $M \gets M_g(idx)$, $\mat{N} \gets \mat{N_g}(idx)$, $D \gets D_g(idx)$
    \label{alg:reduce_scatter:shard}

    \For{$i \gets 2$ to $\rootp$}
    \State $M, \mat{N}, D \gets$ \fn{SendRecv}($\set{M, \mat{N}, D}$, $p_{r,l}$, $p_{r,g}$)
    \label{alg:reduce_scatter:sendrecv}\Comment{Send to left, receive from right}
    \State $idx \gets \set{(c+i)\%\rootp + j\rootp~|~j\in [0,N/\rootp)}$ \Comment{Update the
    indices}\label{alg:reduce_scatter:new_idx}

    \State $M, \mat{N}, D \gets$ \attnfix($M_g(idx), \mat{N_g}(idx), D_g(idx), M, \mat{N}, D$)
    \Comment{Reduce using Alg.~\ref{alg:reduce_op}} \label{alg:reduce_scatter:reduce_op}
    \EndFor
    \State Return $M, \mat{N}, D$\label{alg:reduce_scatter:return}
\end{algorithmic}
    \caption{\label{alg:reduce_scatter}\fn{RowReduceScatter}: Ring reduce-scatter within rows of
    processor grid. The resulting output matrix will be 1D cyclically
    distributed with partitioning along the sequence length dimension.}
\end{algorithm}

\section{\label{sec:tiled_attn_bwd}2D Parallel Self-attention Backward Pass With Overlapping
Communication}
In this section, we explain the backward pass of a 2D parallel self-attention operation that
involves overlapping computation and communication.
Alg.~\ref{alg:attn_bwd} provides the pseudocode for this backward pass. The computational flow is
quite similar to the forward pass described in Alg.~\ref{alg:attn_fwd}.
As in the forward pass, the asynchronous send and receive of the first block of the query matrix
$\mat{Q_p}$ is initiated in line~\ref{alg:attn_bwd:sendrecv}.
Concurrently, the function \gcbwd (Alg.~\ref{alg:gcbwd}) is called in line~\ref{alg:attn_bwd:gc}
to gather the key and value matrices within a column of the processor grid, while overlapping this
communication with the computation of backward pass of self-attention for the query block $\mat{Q_p}$. The function
\gcbwd is very similar to \gc, except that \gcbwd calls the function \flashattnbwd
(in line~\ref{alg:gcbwd:compute} of Alg.~\ref{alg:gcbwd})
to compute the gradients of the query, key and value blocks, and uses elementwise summation for local
reduction (in line~\ref{alg:gcbwd:reduce} of Alg.~\ref{alg:gcbwd}).

\begin{algorithm}
\begin{algorithmic}[1]
    \Require
    $\mat{Q_p}, \mat{O_p}, \mat{dO_p} \in \mathbb{R}^{(N/P) \times H}$: Query, output, and output
    gradient matrices cyclically distributed in column-major fashion to a processor $p_{r,c}$;\\
    $\mat{K_t}, \mat{V_t} \in \mathbb{R}^{(N/P) \times H}$: Key and value matrices cyclically
    distributed in row-major fashion to a processor $p_{r,c}$;\\
    $M_p, D_p \in \rspace{N/P}$: Softmax statistics computed during forward pass cyclically
    distributed in column-major fashion to a processor $p_{r,c}$;\\
    $p_{r,c}$: Current processor; $P$: Total number of processors

    \State $l \gets (c + \rootp - 1) \% \rootp$, $g \gets (c+1)\%\rootp$
    \Comment{Left and right processor column ids}

    \State $\mat{Q_n}, \mat{O_n}, \mat{dO_n}, M_n, D_n, h_g$ $\gets$
    \fn{AsyncSendRecv}($\set{\mat{Q_p}, \mat{O_p}, \mat{dO_p}, M_p, D_p}$, $p_{r,l}$, $p_{r,g}$)
    \label{alg:attn_bwd:sendrecv}
    \State $\mat{K_g}, \mat{V_g}, \mat{dQ_1}, \mat{dK_g}, \mat{dV_g} \gets \gcbwd(\mat{Q_p}$,
    $\mat{K_t}$, $\mat{V_t}$, $\mat{O_p}$, $\mat{dO_p}$, $M_p$, $D_p$, $p_{r,l}$, $p_{r,g}$, $P)$
    \label{alg:attn_bwd:gc} \Comment{Alg.~\ref{alg:gcsbwd}}

    \State $\fn{Wait}(h_g)$\label{alg:attn_bwd:wait}
    \State $\mat{dQ_2}, \mat{dK_p}, \mat{dV_p} \gets \gcsbwd(\mat{Q_n}$, $\mat{K_g}$,
    $\mat{V_g}$, $\mat{O_n}$, $\mat{dK_g}$, $\mat{dV_g}$, $\mat{dO_n}$, $M_n$, $D_n$, $p_{r, l}$,
    $p_{r, g}$, $P)$

    \State $\mat{dQ_p} \gets \mat{dQ_1} + \mat{dQ_2}$
    \State \Return $\mat{dQ_p}, \mat{dK_p}, \mat{dV_p}$
\end{algorithmic}
    \caption{\label{alg:attn_bwd}\fn{\attnbwd}: Backward pass for \attn.}
\end{algorithm}

\begin{algorithm}
\begin{algorithmic}[1]
    \Require
    $\mat{Q_p}, \mat{K_1}, \mat{V_1}, \mat{O_p}, \mat{dO_p} \in \mathbb{R}^{(N/P)\times H}$: Query,
    key, value, output and gradient matrices cyclically distributed to a processor $p_{r,c}$;\\
    $M_p, D_p \in \rspace{N/P}$: Softmax statistics computed during forward pass cyclically
    distributed in column-major fashion to a processor $p_{r,c}$;\\
    $p_{r,c}$: Current processor; $P$: Total number of processors

    \State $u \gets (r + \rootp - 1) \% \rootp$, $d \gets (r+1)\%\rootp$
    \Comment{Up and down processor row ids}
    \For{$i \gets 1$ to $\rootp$}
    \IfThen{$i > 1$}{$\fn{Wait}(h_g)$} \Comment{Wait for gather data to arrive}
    \IfThen{$i < \rootp$}{$\mat{K_{i+1}}, \mat{V_{i+1}}, h_g \gets
    \fn{AsyncSendRecv}(\set{\mat{K_i}, \mat{V_i}}, p_{u,c}, p_{d,c})$}
    \label{alg:gcbwd:gather} \Comment{Gather}

    \State $\mat{dQ_i}, \mat{dK_i}, \mat{dV_i} \gets \flashattnbwd(\mat{Q_p}$, $\mat{K_i}$,
    $\mat{V_i}$, $\mat{O_p}$, $\mat{dO_p}$, $M_p$, $D_p$)
    \label{alg:gcbwd:compute}\Comment{Compute}
    \If{$i > 1$}
    \State $\mat{dQ} \gets \mat{dQ} + \mat{dQ_i}$
    \label{alg:gcbwd:reduce}\Comment{Reduce}
    \Else
    \State $\mat{dQ} \gets \mat{dQ_i}$
    \EndIf
    \EndFor
    \State \Return $\fn{Concat}(\mat{K_1},\dots,\mat{K_{\rootp}})$,
    $\fn{Concat}(\mat{V_1},\dots,\mat{V_{\rootp}})$, $\mat{dQ}$,
    $\fn{Concat}(\mat{dK_1},\dots,\mat{dK_{\rootp}})$,
    $\fn{Concat}(\mat{dV_1},\dots,\mat{dV_{\rootp}})$
\end{algorithmic}
    \caption{\label{alg:gcbwd}\gcbwd: Overlapping all-gather and compute operation (with local reduction)
    for backward pass that overlaps computation and communication.}
\end{algorithm}

Once the next query block is received (line~\ref{alg:attn_bwd:wait} of Alg.~\ref{alg:attn_bwd}), the
remaining gradient computation is carried out using the function \gcsbwd (Alg.~\ref{alg:gcsbwd}),
where the computation is overlapped with gathering the remaining query blocks and reduce-scattering
the partial outputs.  The function \gcsbwd is similar to \gcs, with one key difference: during the
final iteration, in addition to reduce-scattering the query gradients within the processor rows, the
key and value gradients must also be reduce-scattered within the processor columns. This is achieved
by calling the function \csbwd in line~\ref{alg:gcsbwd:csbwd} of Alg.~\ref{alg:gcsbwd}.  The
pseudocode for \csbwd is described in Alg.~\ref{alg:csbwd}. \csbwd computes self-attention gradient
for the final query block $\mat{Q_n}$ and the key and value matrices $\mat{K_g}$ and $\mat{V_g}$ on
a block-by-block basis (as shown in line~\ref{alg:csbwd:compute} of Alg.~\ref{alg:csbwd}), while
simultaneously overlapping this computation with the reduce-scatter communication of key and value
gradients (as shown in line~\ref{alg:csbwd:reduce} of Alg.~\ref{alg:csbwd}).

\begin{algorithm}
\begin{algorithmic}[1]
    \Require
    $\mat{Q_n}, \mat{K_g}, \mat{V_g}, \mat{O_n}, \mat{dK_g}, \mat{dV_g}, \mat{dO_n}$: Query,
    key, value, output and gradient matrices cyclically distributed to a processor $p_{r,c}$;\\
    $M_n, D_n \in \rspace{N/P}$: Softmax statistics computed during forward pass cyclically
    distributed in column-major fashion to a processor $p_{r,c}$;\\
    $p_{r,c}$: Current processor; $P$: Total number of processors

    \State $l \gets (c + \rootp - 1) \% \rootp$, $g \gets (c+1)\%\rootp$
    \Comment{Left and right processor column ids}

    \For{$i \gets 1$ to $\rootp-1$}
    \IfThen{$i > 1$}{$\fn{Wait}(h_g)$} \Comment{Wait for gather data to arrive}
    \If{$i < \rootp-1$}
    \State $\mat{Q_b}$, $\mat{O_b}$, $\mat{dO_b}$, $M_b$, $D_b$, $h_g \gets$
    \fn{AsyncSendRecv}($\set{\mat{Q_n}, \mat{O_n}, \mat{dO_n}, M_n, D_n}$, $p_{r,l}$, $p_{r,g}$)
    \label{alg:gcsbwd:gather} \Comment{Gather}
    \State $\mat{dQ_p}, \mat{dK_p}, \mat{dV_p} \gets \flashattnbwd(\mat{Q_n}$, $\mat{K_g}$,
    $\mat{V_g}$, $\mat{O_n}$, $\mat{dO_n}$, $M_n$, $D_n$)
    \label{alg:gcsbwd:compute} \Comment{Compute}
    \State $\mat{dK_g} \gets \mat{dK_g} + \mat{dK_p}$, $\mat{dV_g} \gets \mat{dV_g} + \mat{dV_p}$
    \Comment{Reduce $\mat{dK}$ and $\mat{dV}$}
    \Else
    \State $\mat{dQ_p}, \mat{dK}, \mat{dV} \gets \csbwd(\mat{Q_n}$, $\mat{K_g}$,
    $\mat{V_g}$, $\mat{O_n}$, $\mat{dO_n}$, $M_n$, $D_n$)
    \label{alg:gcsbwd:csbwd}
    \Comment{Overlapping Compute and ReduceScatter of $\mat{dK_g}$ and $\mat{dV_g}$}
    \EndIf

    \If{$i > 1$}
    \State $\fn{Wait}(h_s)$ \label{alg:gcsbwd:wait_hs} \Comment{Wait for previous scatter data to arrive}
    \State $\mat{dQ_p} \gets \mat{dQ_p} + \mat{dQ}$ \label{alg:gcsbwd:reduce}
    \Comment{Reduce $\mat{dQ}$}
    \EndIf

    \State $\mat{dQ} \gets \fn{AsyncSendRecv}(\mat{dQ_p}, p_{r,l}, p_{r,g})$
    \label{alg:gcsbwd:scatter} \Comment{Scatter}

    \State $\mat{Q_n}$, $\mat{O_n}$, $\mat{dO_n}$, $\mat{Q_b}$, $\mat{O_b}$, $\mat{dO_b} \gets
    \mat{Q_b}$, $\mat{O_b}$, $\mat{dO_b}$, $\mat{Q_n}$, $\mat{O_n}$, $\mat{dO_n}$ \Comment{Swap
    buffer pointers}
    \EndFor

    \State $\fn{Wait}(h_s)$ \Comment{Wait for final scatter data to arrive}
    \State \Return $\mat{dQ}$, $\mat{dK}$, $\mat{dV}$
\end{algorithmic}
    \caption{\label{alg:gcsbwd}\gcsbwd: Overlapping all-gather, compute and reduce-scatter operation for
    backward pass that overlaps computation and communication.}
\end{algorithm}

\begin{algorithm}
\begin{algorithmic}[1]
    \Require
    $\mat{Q_n}, \mat{K_g}, \mat{V_g}, \mat{O_n}, \mat{dK_g}, \mat{dV_g}, \mat{dO_n}$: Query,
    key, value, output and gradient matrices cyclically distributed to a processor $p_{r,c}$;\\
    $M_n, D_n \in \rspace{N/P}$: Softmax statistics computed during forward pass cyclically
    distributed in column-major fashion to a processor $p_{r,c}$;\\
    $p_{r,c}$: Current processor; $P$: Total number of processors

    \State $u \gets (r + \rootp - 1) \% \rootp$, $d \gets (r+1)\%\rootp$
    \Comment{Up and down processor row ids}
    \State $\mat{K_1},\dots,\mat{K_{\rootp}} \gets \fn{Split}(\mat{K_g})$
    \Comment{Split along sequence dim with cyclic distribution}
    \State $\mat{V_1},\dots,\mat{V_{\rootp}} \gets \fn{Split}(\mat{V_g})$
    \State $\mat{dK_1},\dots,\mat{dK_{\rootp}} \gets \fn{Split}(\mat{dK_g})$
    \State $\mat{dV_1},\dots,\mat{dV_{\rootp}} \gets \fn{Split}(\mat{dV_g})$

    \For{$i \gets 2$ to $\rootp$}
    \State $\mat{dQ_p}, \mat{dK_p}, \mat{dV_p} \gets \flashattnbwd(\mat{Q_n}$, $\mat{K_i}$,
    $\mat{V_i}$, $\mat{O_n}$, $\mat{dO_n}$, $M_n$, $D_n$) \Comment{Compute}
    \label{alg:csbwd:compute}
    \State $\mat{dK_p} \gets \mat{dK_p} + \mat{dK_i}, \mat{dV_p} \gets \mat{dV_p} + \mat{dV_i}$
    \If{$i > 2$}
    \State $\mat{dQ} \gets \mat{dQ} + \mat{dQ_p}$ 
    \State \fn{Wait}($h_s$)
    \State $\mat{dK_p} \gets \mat{dK_p} + \mat{dK}, \mat{dV_p} \gets \mat{dV_p} + \mat{dV}$
    \Else
    \State $\mat{dQ} \gets \mat{dQ_p}$
    \EndIf
    \State $\mat{dK}, \mat{dV}, h_s \gets \fn{AsyncSendRecv}(\set{\mat{dK_p}, \mat{dV_p}}, p_{u,c},
    p_{d,c})$ \label{alg:csbwd:scatter} \Comment{Scatter} \label{alg:csbwd:reduce}
    \EndFor
    \State $\mat{dQ_p}, \mat{dK_p}, \mat{dV_p} \gets \flashattnbwd(\mat{Q_n}$, $\mat{K_1}$,
    $\mat{V_1}$, $\mat{O_n}$, $\mat{dO_n}$, $M_n$, $D_n$)
    \State $\mat{dQ} \gets \mat{dQ} + \mat{dQ_p}$ 
    \State \fn{Wait}($h_s$)
    \State $\mat{dK} \gets \mat{dK_p} + \mat{dK}, \mat{dV} \gets \mat{dV_p} + \mat{dV}$
    \State \Return $\mat{dQ}, \mat{dK}, \mat{dV}$
\end{algorithmic}
    \caption{\label{alg:csbwd}\csbwd: Overlapping compute, and reduce-scatter. The resulting output matrix
    will be 1D cyclically distributed with partitioning along the sequence length dimension.}
\end{algorithm}

\end{document}